\documentclass[10pt, twocolumn, letterpaper]{article}

\usepackage[pagenumbers]{cvpr} 

\usepackage{graphicx}
\usepackage{amsmath}
\usepackage{amssymb}
\usepackage{booktabs}
\usepackage{verbatim}

\usepackage{multirow}
\usepackage[dvipsnames, table, xcdraw]{xcolor}
\usepackage{pifont}
\newcommand{\cmark}{\ding{51}}%

\newcommand{\figref}[1]{Fig.~\ref{#1}}
\newcommand{\tabref}[1]{Tab.~\ref{#1}}
\newcommand{\eqnref}[1]{Eqn.~\ref{#1}}

\newcommand{\myPara}[1]{\vspace{5pt}\noindent\textbf{#1}}
\newcommand{\sArt}{SOTA~}

\graphicspath{{./figs/}}

\usepackage{makecell}
\newcommand{\highlight}[1]{\textbf{\textcolor{BrickRed}{#1}}}

\newcommand{\addFig}[1]{}
\newcommand{\addFigs}[1]{}

\usepackage[pagebackref, breaklinks, colorlinks, linkcolor=BrickRed, citecolor=RoyalBlue]{hyperref}

\usepackage[capitalize]{cleveref}
\crefname{section}{Sec.}{Secs.}
\Crefname{section}{Section}{Sections}
\Crefname{table}{Table}{Tables}
\crefname{table}{Tab.}{Tabs.}

\def\cvprPaperID{1144} 
\def\confName{CVPR}
\def\confYear{2023}

\newcommand{\nameofmethod}{CamoFormer}

\begin{document}

\title{\nameofmethod{}: Masked Separable Attention for Camouflaged Object Detection}

\author{Bowen Yin$^1$ \quad Xuying Zhang$^1$ \quad Qibin Hou$^{1*}$ 
  \quad Bo-Yuan Sun$^1$ \quad Deng-Ping Fan$^2$ \quad Luc Van Gool$^2$  \\ \\
$^1$ TMCC, School of Computer Science, Nankai University \\
$^2$ ETH Zurich\\
{\tt\small https://github.com/HVision-NKU/CamoFormer}
}


\twocolumn[{%
\renewcommand\twocolumn[1][]{#1}%
\maketitle
\vspace{-15pt}
\begin{center}
    \centering
    \captionsetup{type=figure}   \includegraphics[width=\textwidth,height=.2\textwidth]{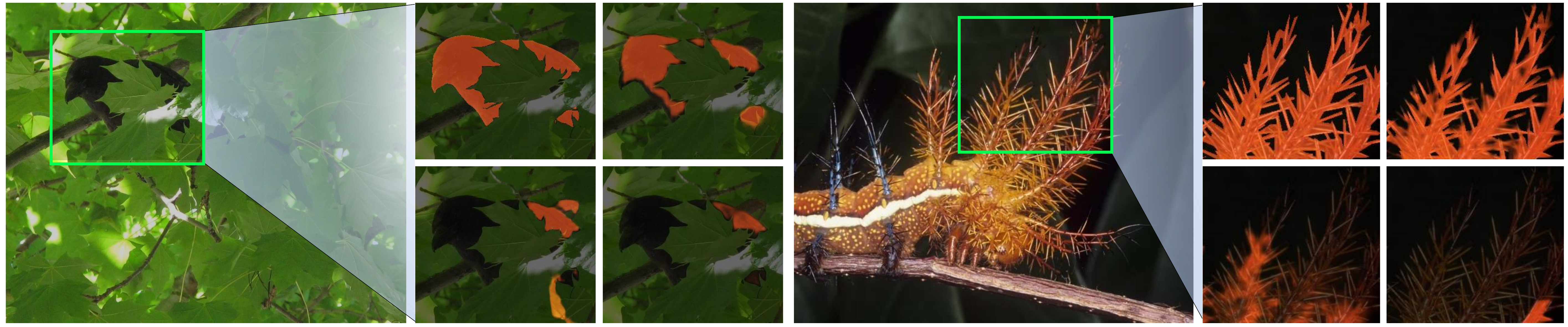}
    \put (-362, 53){\footnotesize{{\color{white}{GT}}}}
    \put (-112, 53){\footnotesize{{\color{white}{GT}}}}
    \put (-302, 53){\footnotesize{{\color{white}{Ours}}}}
    \put (-53, 53){\footnotesize{{\color{white}{Ours}}}}
    \put (-362, 4){\footnotesize{{\color{white}{SegMaR}}}}
    \put (-112, 4){\footnotesize{{\color{white}{SegMaR}}}}
    \put (-302, 3){\footnotesize{{\color{white}{ZoomNet}}}}
    \put (-53, 3){\footnotesize{{\color{white}{ZoomNet}}}}
    \vspace{-5pt}
    \captionof{figure}{
    \small Visual comparison between our \nameofmethod{} and recent state-of-the-art methods (\eg, SegMaR~\cite{Jia_2022_CVPR} and ZoomNet~\cite{Pang_2022_CVPR}) for camouflaged object detection. 
    The segmentation details of different methods in the green rectangle regions are displayed with focus views. We can easily observe that our \nameofmethod{} can generate much better results than other methods. Best viewed in color.}
    \label{fig:main_comparison}
\end{center}%
}]

\begin{abstract}
%
How to identify and segment camouflaged objects from the background is challenging.
Inspired by the multi-head self-attention in Transformers, we present a simple masked separable attention (MSA) for camouflaged object detection.
%
We first separate the multi-head self-attention into three parts, which are responsible for distinguishing the camouflaged objects from the background using different mask strategies.
Furthermore, we propose to capture high-resolution semantic representations progressively based on a simple top-down decoder with the proposed MSA
to attain precise segmentation results.
These structures plus a backbone encoder
form a new model, dubbed \textbf{\nameofmethod}. 
Extensive experiments show that \nameofmethod{} surpasses all existing state-of-the-art methods on three widely-used camouflaged object detection benchmarks. 
There are on average $\sim$\textbf{5\%} relative improvements over previous methods
in terms of S-measure and weighted F-measure.
\end{abstract}

\let\thefootnote\relax\footnotetext{$^{*}$Qibin Hou is the corresponding author.}

\vspace{-15pt}
\section{Introduction}
\label{sec:intro}
Camouflaged object detection (COD) is a new challenging task~\cite{fan2022concealed} that has been popular in recent years~\cite{fan2020camouflaged,fan2020pranet,mei2021camouflaged,fan2022concealed,ji2022deep}.
%
Biological studies have shown that the human visual perceptual system can be easily
deceived~\cite{stevens2009animal} by various camouflage strategies in that 
camouflaged objects are highly similar to their surroundings or extremely small 
in size.
The high similarity between the camouflaged objects and their surroundings 
makes the COD~\cite{fan2020camouflaged} more challenging than traditional object detection~\cite{medioni2009generic,liu2020deep}, attracting more and more research attention, such as medical image segmentation~\cite{fan2020pranet} and search engine~\cite{fan2022concealed}.

%
%

%

There are an increasing number of works using sophisticated 
deep learning techniques~\cite{mei2021camouflaged, Pang_2022_CVPR, Zhong_2022_CVPR} to solve this task, especially after a large-scale dataset was proposed~\cite{fan2022concealed}.
%
However, even the state-of-the-art (SOTA) methods are still struggled to segment camouflaged targets with fine shapes for some complex scenes, as shown in~\figref{fig:main_comparison}.
We argue that an important reason is that these models cope with the foreground and background cues indiscriminately, making them difficult to identify the camouflaged objects from similar surroundings.
%
The key to solving this issue is to encode the foreground and background cues separately. 

Taking the aforementioned analysis into account, in this paper, we present \emph{Masked Separable Attention (MSA)} to explicitly process the camouflaged objects
and background based on the Multi-Dconv Head Transposed Attention~\cite{zamir2022restormer}. 
Instead of performing the same operations in each head,
our MSA separates the attention heads into three groups, 
each of which is responsible for specific functionality.
In particular, inspired by masked attention~\cite{cheng2022masked}, we propose to use two groups of heads to process the foreground and background regions independently.
Our goal is to use the attention scores built within the predicted foreground generated by a prediction head to discover camouflaged objects from the full-value representations and vice versa for the background.
%
%
Besides, we keep a group of normal attention heads for building global interactions, which is essential for generating high-quality segmentation maps.

Given the proposed MSA, we apply it to an encoder-decoder architecture~\cite{lin2017feature, liu2019simple} to 
progressively refine the segmentation map as illustrated in~\figref{fig:network}.
At each feature level of the decoder, a segmentation map is predicted
and sent to an MSA block to improve the prediction
quality.
This progressive refinement process enables us to attain high-quality camouflaged object predictions as the feature resolution increases.
As shown in~\figref{fig:main_comparison}, 
our \nameofmethod{} can more accurately identify the camouflaged objects
and generate segmentation maps with finer borders than other cutting-edge methods.

To validate the effectiveness of our \nameofmethod, we conduct extensive experiments on three popular COD benchmarks (NC4K~\cite{lv2021simultaneously}, COD10K~\cite{fan2020camouflaged}, and CAMO~\cite{le2019anabranch}). 
On all these benchmarks, \nameofmethod{} achieves new state-of-the-art (SOTA) records compared to the recent methods.  
In particular, our method achieves 0.786 weighted F-measure and 0.023 MAE, while the corresponding results for the second-best model FDNet~\cite{Zhong_2022_CVPR} are 0.731 and 0.030 on the COD10K-test dataset.
Furthermore, the visualization results also show the superiority of our \nameofmethod~over existing COD methods.
To sum up, our main contributions can be summarized as follows:
\begin{itemize}
    \item We present masked separable attention (MSA).
    It divides the attention heads into three groups and discovers the
    foreground and background regions by separately computing their attention scores via a predicted map for accurate segmentation; 
    \item We adopt a progressive refinement manner to make better use of our MSA, which can gradually improve the segmentation results predicted at each feature level;
    \item Our \nameofmethod{} improves all the recent \sArt methods by a large margin on all these three COD benchmarks. 
    Visualization experiments further illustrate the superiority of \nameofmethod{} on the completeness of the segments. 
\end{itemize}


\section{Related Work}
\label{sec:formatting}

\subsection{Camouflaged Object Detection}
%

Traditional COD methods~\cite{zhang2016bayesian, beiderman2010optical, galun2003texture, guo2008robust, kavitha2011efficient, hall2013camouflage} extract various hand-crafted features between the camouflaged objects and backgrounds to segment the camouflaged targets.  
These methods can deal with simple scenes, but show drastic performance degradation in complex conditions. 

Recently, the mainstream in COD is CNN-based approaches~\cite{ji2022fast, Zhong_2022_CVPR, Pang_2022_CVPR, zhang2022preynet,Jia_2022_CVPR, mei2021camouflaged, fan2020camouflaged}, which can be categorized into three strategies: 
i) multi-scale feature aggregation: 
CubeNet~\cite{zhuge2022cubenet} accompanies attention fusion and X-shaped connection to integrate features from multiple layers sufficiently. 
ZoomNet~\cite{Pang_2022_CVPR} processes the input images at three scales and unifies the scale-specific appearance features at different scales. 
%
%
ii) Multi-stage strategy: 
Due to the concealment of camouflaged objects, SINet~\cite{fan2020camouflaged} proposes first to locate and then distinguish them for better performance. 
PreyNet~\cite{zhang2022preynet} mimics the process of predation and splits the detection process of camouflaged targets into initial detection and predator learning.
SINetV2~\cite{fan2022concealed} adopts surrounding connection decoder and group-reversal attention to improve the performance. 
SegMaR~\cite{Jia_2022_CVPR}, a multi-stage training and inference framework, locates the target and magnifies the object regions to detect camouflaged objects progressively.  
iii) Joint training strategy: 
UJSC~\cite{li2021uncertainty} leverages the contradictory information to enhance the detection ability for both salient object detection and camouflaged object detection.

%

\begin{figure*}[htbp]
\centering
\includegraphics[width=\linewidth]{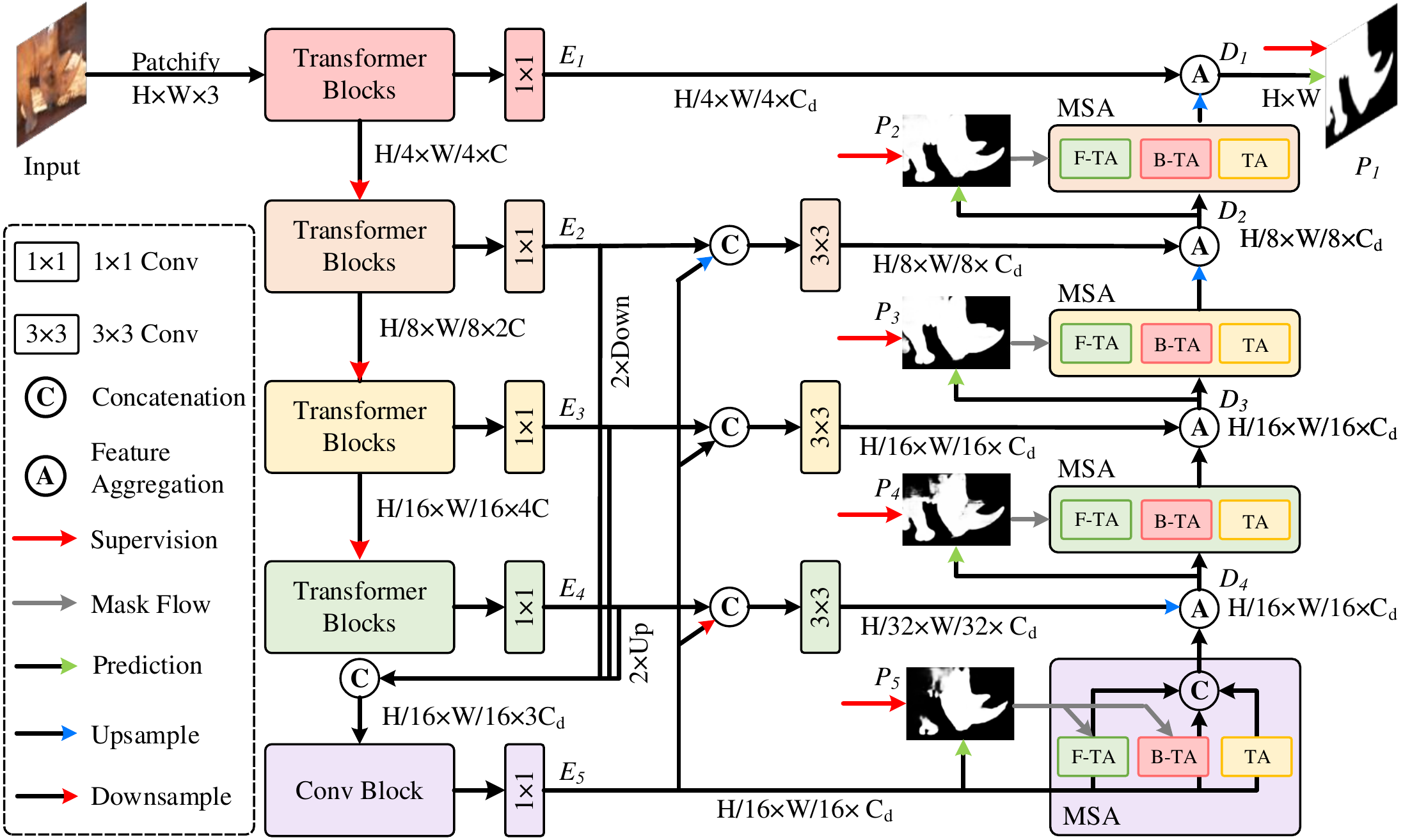}
\vspace{-5pt}
\caption{Overall architecture of our \nameofmethod{} model. 
First, a pretrained Transformer-based backbone is utilized to extract multi-scale features of the input image. Then, the features from the last three stages are aggregated to generate the coarse prediction. Next, the progressive refinement decoder equipped with masked separable attention (MSA) is applied to gradually polish the prediction results.
All the predictions generated by our \nameofmethod~are supervised by the ground truth (GT).
}\label{fig:network}
\vspace{-5pt}
\end{figure*}

\subsection{Transformers in Computer Vision}
Transformer architectures are initially designed for natural language processing~\cite{vaswani2017attention, devlin2018bert} and then get popular in computer vision~\cite{srinivas2021bottleneck,wu2021cvt,tnt,xu2021coat,chu2021twins,yang2021focal}.
%
Compared with conventional convolutional neural networks~\cite{he2016deep,simonyan2014very,szegedy2016rethinking,hu2018squeeze,regnet},
Transformers can efficiently encode global contextual information and hence 
have been widely used in a variety of visual tasks, including image classification~\cite{dosovitskiy2020image, touvron2021training, yuan2022volo, wang2021pyramid}, semantic segmentation~\cite{xie2021segformer, zheng2021rethinking,jiang2021all}, object detection~\cite{carion2020end},
and salient object detection~\cite{zhuge2022salient,liu2021visual}.
%

Transformer-based models are also becoming a new trend in COD.
UGTR~\cite{yang2021uncertainty} explicitly utilizes the probabilistic representational model to learn the uncertainties of the camouflaged object under the Transformer framework. 
DTINet~\cite{liu2022boosting} designs a dual-task interactive Transformer to segment both the camouflaged objects and their detailed borders. 
TPRNet~\cite{zhang2022tprnet} proposes a transformer-induced progressive refinement network that utilizes the semantic information from high-level features to guide the detection of camouflaged targets. 


Our \nameofmethod{} is also built upon the popular Transformer framework.
Not focusing on a novel architecture design, we aim to investigate more efficient ways to utilize self-attention for COD and receive better performance than other methods.
We assign different functionalities to different attention heads to process the foreground and background regions separately, which makes our work quite different from other Transformer-based COD methods.


\section{Proposed \nameofmethod{}}

Similar to most previous works~\cite{fan2020camouflaged,fan2020pranet,Pang_2022_CVPR,Zhong_2022_CVPR,jiang2022magnet}, we adopt an encoder-decoder architecture
to build our \nameofmethod{}. 
%
Our \nameofmethod{} is an end-to-end trainable framework, which is shown in \figref{fig:network}. 
%

%

%


\subsection{Overall Architecture}
\myPara{Encoder.} By default, we adopt the PVTv2~\cite{wang2022pvt} as our encoder, as vision transformers have shown great performance in binary segmentation tasks~\cite{liu2021visual, zhang2021learning}.
%
Given an input image $I\in\mathbb{R}^{H \times W \times 3}$, we feed it into the encoder to generate multi-scale feature maps from the four stages, which are denoted as $\left \{ E_i\right \}_{i=1}^4$.
Consequently, $E_1$ is with spatial size $\frac{H}{4} \times \frac{W}{4}$ and 
$E_4$ is with spatial size $\frac{H}{32} \times \frac{W}{32}$.
Then, we aggregate the features from the last three stages of the encoder and
send them to a convolutional block, yielding representations $E_5$ with higher-level semantics. 
%

\myPara{Decoder.}
The decoder is built upon the encoder.
The multi-level semantic features $\left \{ E_i\right \}_{i=1}^5$ from the encoder are fed into the decoder.
To achieve a better trade-off between efficiency and performance, we first connect a $1\times1$ convolution with $C_d = 128$ channels to the feature maps at each level.
As shown in~\figref{fig:network}, we adopt a progressive way to refine the features
from the top of the encoder.
At each feature level, masked separable attention (MSA) is used for a better
distinguishment of the camouflaged objects and the background.
%
In the initial level of progressive fusion, the aggregated feature $D_4$ can be written as:
\begin{equation}
   D_{4} = \mathrm{MSA} (E_{5}) \cdot \mathcal{F}\mathrm{_{up}} (E_{4}) + \mathcal{F}\mathrm{_{up}} (E_{4}), 
  \label{eq:ITF0}
\end{equation}
where  $\mathcal{F}\mathrm{_{up}} (\cdot)$ is a bilinear upsampling operation for shape matching. 
And the aggregated features $\left \{ D_i\right \}_{i=1}^{3}$ in the following levels can be defined as:
\begin{equation}
    D_{i} = \mathcal{F}\mathrm{_{up}} (\mathrm{MSA} (D_{i+1})) \cdot E_{i}+ E_{i}. 
  \label{eq:ITF1}
\end{equation}
Unlike previous works~\cite{fan2020camouflaged,Jia_2022_CVPR,Pang_2022_CVPR} that mainly use the addition operation or
the concatenation operation to fuse the features from different feature levels,
we first compute the element-wise product between them and then use the summation operation.
We empirically found that such a simple modification brings about
0.2\%+ relative improvement in terms of S-measure and weighted F-measure averagely on NC4K~\cite{lv2021simultaneously}, COD10K-test~\cite{fan2022concealed}, and CAMO-test~\cite{le2019anabranch}.

\myPara{Loss Function.}
Following~\cite{hou2019deeply,xie2015holistically}, we add
side supervision at each feature level.
We denote the predictions generated by the decoder of \nameofmethod{} as ${\left \{ P_i \right \}}_{i=1}^{5}$.  
Except for the final prediction map $P_1$, all the other prediction maps $P_i$ 
are used in the MSAs for the progressive refinement as described above. 
%
During training, each $P_i$ is rescaled to the same size as the input image and all of them are supervised by the BCE loss~\cite{de2005tutorial} and IoU loss~\cite{mattyus2017deeproadmapper}.
Following~\cite{fan2022concealed}, the overall loss is a summation of multi-stage loss. 
The total loss of our \nameofmethod{} can be formulated as follows:

\begin{equation} 
   \mathcal{L} (P, G) = \sum_{i=1}^5 \mathcal{L}_{bce} (P_i, G)+\mathcal{L}_{iou} (P_i, G), 
  \label{eq:loss}
\end{equation}
where $G$ is the ground truth annotation. 

\subsection{Masked Separable Attention} \label{sec:MSAM}

Camouflaged objects are diverse in scale and highly similar to the background, which makes them difficult to segment completely.
How to accurately identify camouflaged objects from the background is crucial.
We solve this by presenting masked separable attention (MSA), where different
attention heads take charge of different functionalities.
Our intention is to use part of the attention heads to separately calculate the attention scores in the predicted foreground and background regions and use them to identify the camouflaged objects better. 


Our MSA is based on a modified version of self-attention to save computations, namely Multi-Dconv Head Transposed Attention~\cite{zamir2022restormer}, which we denote as TA for short.
Given an input $\mathbf{X}\in\mathbb{R}^{HW \times C}$ where $H$ and $W$ are respectively the height and width while $C$ is the channel number, the TA is formulated as: 
\begin{equation} 
    \mathrm{TA} (\mathbf{Q}, \mathbf{K}, \mathbf{V})=\mathbf{V}\cdot \mathrm{Softmax} (\frac{\mathbf{Q}^\top\mathbf{K}}{\alpha}),
    \label{eq:ta}
\end{equation}
where $\mathbf{Q}$, $\mathbf{K}$, $\mathbf{V}$ are the query, key, value matrices that can be generated by using three separate $1\times1$ convolutions followed
by a $3\times 3$ depthwise convolution, and $\alpha$ is a learnable scaling parameter. 
In practical use, \eqnref{eq:ta} can also be extended to a multi-head
version, as done in the original self-attention~\cite{vaswani2017attention}, 
to augment the feature representations.

\begin{figure}[tp]
  \setlength{\abovecaptionskip}{0.6cm}
  \centering
  \includegraphics[width=0.99\linewidth]{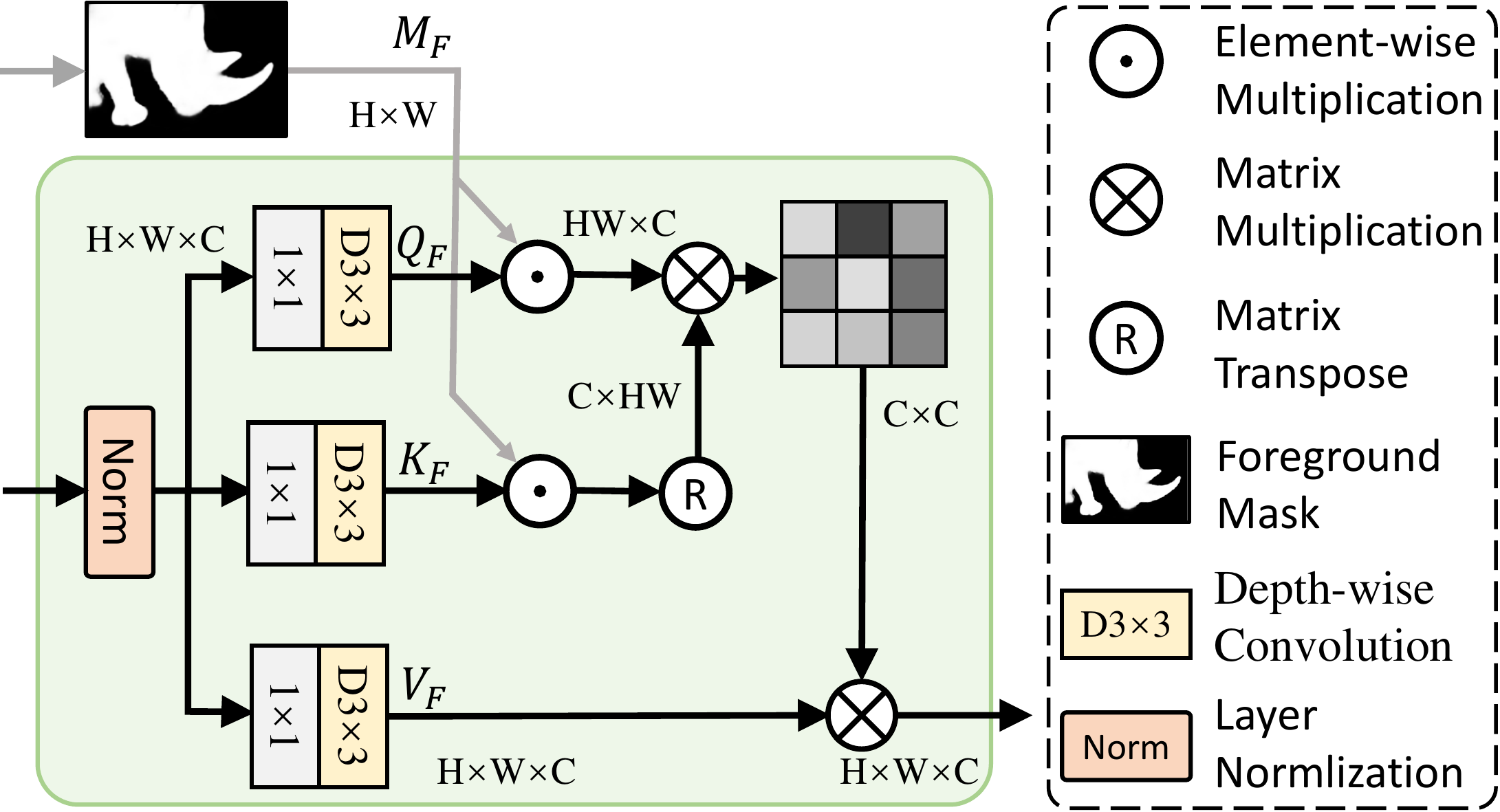}
  \vspace{-8pt}
  \caption{Diagrammatic details of the proposed F-TA in our MSA. Our B-TA shares a similar structure except for the mask.}
  \vspace{-5pt}
  \label{fig:msa_vis}
\end{figure}

\myPara{Masked Separable Attention.}
The attention heads in the above TA are equally utilized for encoding spatial information.
Differently, in our MSA, we propose to introduce a prediction mask which
can be generated at each feature level into TA as a foreground-background
contrast prior for better recognizing the camouflaged objects.
To achieve this, we divide all the attention heads into three groups:
foreground-head TA (F-TA), background-head TA (B-TA), and the normal TA.
The structural details of our MSA are shown in~\figref{fig:msa_vis}.

To be specific, given a predicted foreground mask $M_F$, the formulation of F-TA can be written as: 
\begin{equation}
    \text{F-TA} (\mathbf{Q}_F, \mathbf{K}_F, \mathbf{V}_F)=\mathbf{V}_F \cdot \mathrm{Softmax} (\frac{\mathbf{Q}_F^\top\mathbf{K}_F}{\alpha_F}),
    \label{eq:f-ta}
\end{equation}
where $\mathbf{Q}_F$, $\mathbf{K}_F$ are the masked
query and key matrices that can be produced by multiplying them
with $M_F$ and $\mathbf{V}_F$ is the value matrix without masking.
In this way, the features can be refined by building pairwise relationships within the foreground regions, avoiding the influence of the background
which may contain contaminative information.
Similarly, given the background mask via the broadcast subtraction $M_B = 1 - M_F$, we can also conduct this process for the background.
Thus, the formulation of B-TA can be written as:
\begin{equation}
    \text{B-TA} (\mathbf{Q}_B, \mathbf{K}_B, \mathbf{V}_B)=\mathbf{V}_B \cdot \mathrm{Softmax} (\frac{\mathbf{Q}_B^\top\mathbf{K}_B}{\alpha_B}).
    \label{eq:b-ta}
\end{equation}

Other than the F-TA heads and B-TA heads, the third group of the heads
is kept unchanged as in \eqnref{eq:ta}, which is used to build relationships
between the foreground and background.
The outputs of all the heads are then concatenated and sent into a $3\times3$
convolution for feature aggregation and map the number of channels to $C_d$:
\begin{equation} 
  \mathbf{Z} = \text{Conv}_{3\times3} ([\text{F-TA}, \text{B-TA}, \text{TA}]),
  \label{eq:fuse}
\end{equation}
where $[\cdots]$ is the concatenation operation.

\myPara{Mask Generation.} 
At each feature level, a mask should be generated by a $3\times3$ convolution following a Sigmoid function and then used in our MSA.
As supervision is added to each feature level, we directly use the predictions
${\{ P_i \}}_{i=2}^{5}$ as masks and sent each of them to the corresponding MSA.
Note that we do not binarize the prediction maps but keep them as continuous maps
ranging from 0 to 1, which we found works better during our experiments.

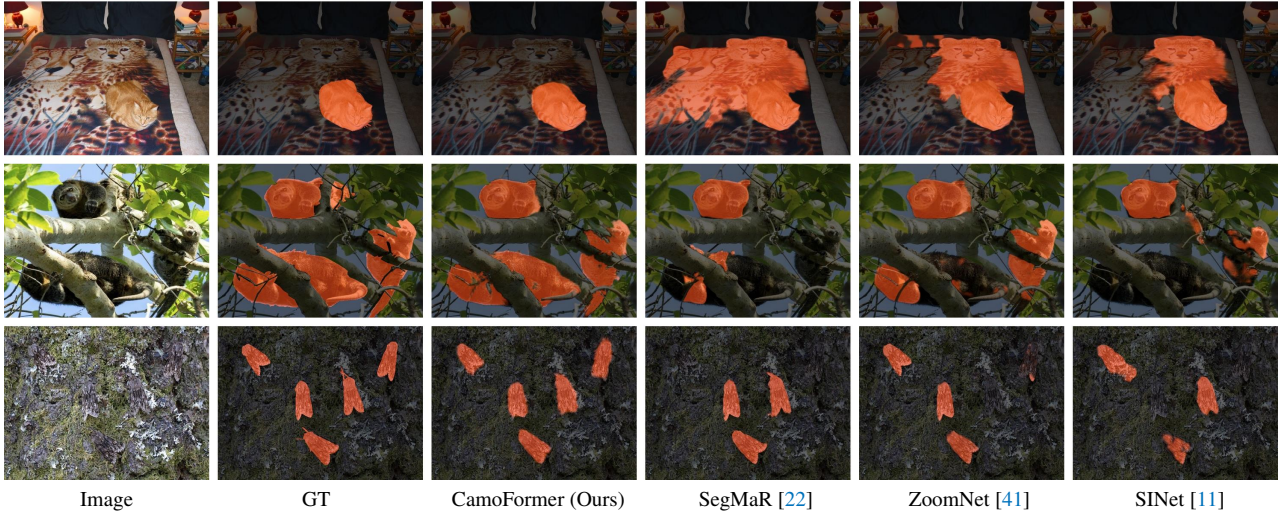
\begin{figure*}[tp]
\setlength{\abovecaptionskip}{0.4cm}
  \centering
  \includegraphics[width=0.97\linewidth]{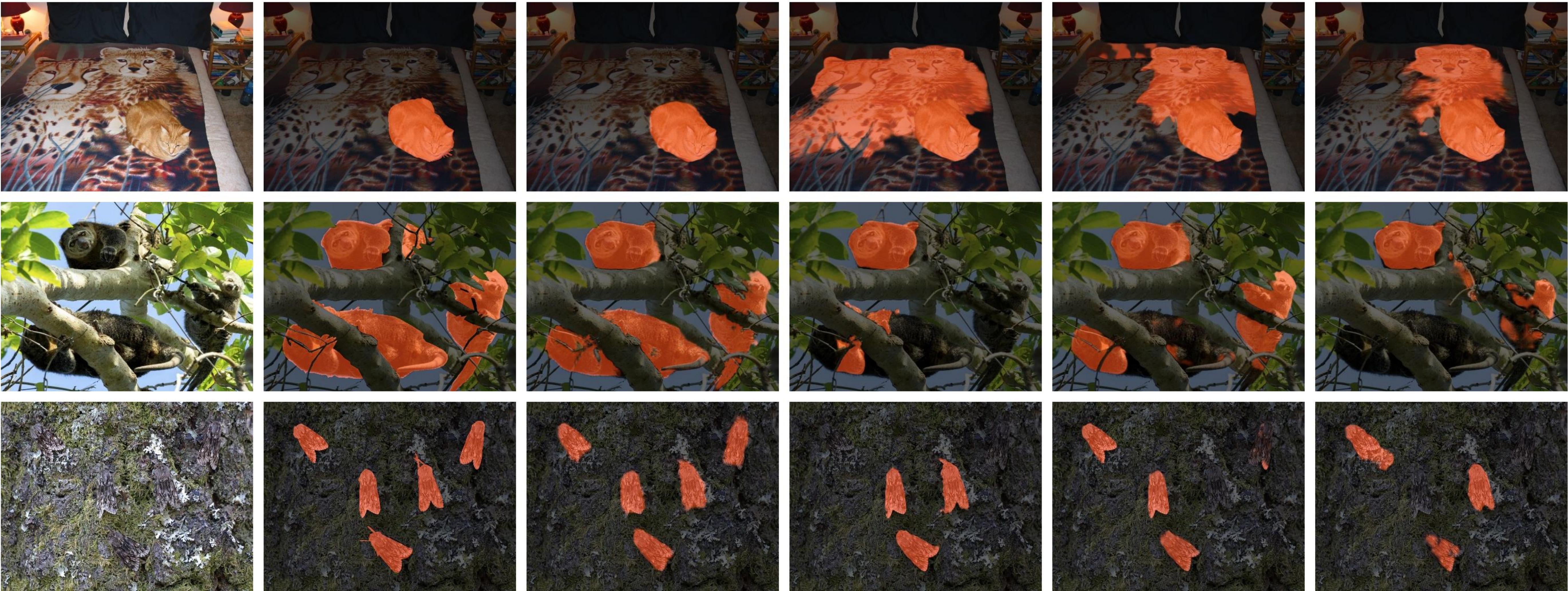}
  \put (-453,-10){\footnotesize{Image}}
  \put (-369,-10){\footnotesize{GT}}
  \put (-314.4,-10){\footnotesize{ \nameofmethod{} (Ours) }}
  \put (-219,-10){\footnotesize{SegMaR}~\cite{Jia_2022_CVPR}}
  \put (-139.6,-10){\footnotesize{ZoomNet}~\cite{Pang_2022_CVPR}}
  \put (-54,-10){\footnotesize{SINet}~\cite{fan2020camouflaged}}
  \vspace{-5pt}
  \caption{Visualization comparisons between our \nameofmethod{} and other SOTA methods. 
  Segmentation results are shown in orange. 
  }\label{fig:vis_method}
\end{figure*}

\section{Experimental Results}\label{sec:experiment}

\subsection{Experiment Setup}
\myPara{Implementation Details.} 
We implement our \nameofmethod{} using the Pytorch library~\cite{paszke2019pytorch}.
A pretrained PVTv2~\cite{wang2022pvt} on the ImageNet dataset~\cite{krizhevsky2017imagenet} is employed as the encoder of our network. 
Unless otherwise specified, we adopt PVTv2~\cite{wang2022pvt} as the backbone.
Besides, we also report results with other backbones, \eg Transformer-based Swin Transformer~\cite{liu2021swin}, CNN-based ResNet~\cite{he2016deep} and ConvNeXt~\cite{liu2022convnet}.
SGD with momentum 0.9 and weight decay 2e-4 is used as the optimizer. 
The learning rate is initially set to 5e-3 and decays following the cosine learning rate strategy.  
During training, all the input images are resized to 384$\times$384. 
The entire model is trained end-to-end for 60 epochs costing around 7 hours with a batch size of 6 on an NVIDIA V100 GPU.  

\myPara{Datasets.}
We evaluate our methods on three popular COD benchmarks, including CAMO~\cite{le2019anabranch}, COD10K~\cite{fan2022concealed}, and NC4k~\cite{lv2021simultaneously}.  
CAMO comprises 2,500 images, half of which contain camouflaged objects and half do not. 
COD10k includes 5,066 camouflaged, 3,000 background, and 1,934 non-camouflaged images. 
NC4K is a large-scale COD dataset consisting of 4,121 images for testing. 
Following previous works~\cite{fan2020camouflaged, Jia_2022_CVPR,Pang_2022_CVPR}, we use 1,000 images from the CAMO dataset and 3,040 images from COD10K for training and the others for testing. 
%

\myPara{Metrics.} Following~\cite{Pang_2022_CVPR, Jia_2022_CVPR, Zhong_2022_CVPR}, we use four golden metrics for evaluation, including Structure-measure (S$_m$)~\cite{fan2017structure}, mean absolute error (M)~\cite{perazzi2012saliency}, weighted F-measure ($w$F)~\cite{margolin2014evaluate}, and adaptive E-measure ($\alpha$E)~\cite{fan2018enhanced}. 
M is the absolute difference between the prediction map and GT. 
S$_m$ simultaneously evaluates region-aware and object-aware
structural similarity between predictions and GT. 
$w$F is an exhaustive measure of both recall and precision. 
$\alpha$E evaluates element-wise similarity and the statistics at the image level. 
In addition, we draw the precision-recall (PR) curves and $F_\beta$-threshold ($F_\beta$) curves in the supplementary materials.

\begin{table*}[tp]
\setlength\tabcolsep{6pt}
\renewcommand{\arraystretch}{0.9}
\small
\centering
\setlength{\belowcaptionskip}{0cm}   
\begin{tabular}{lccccccccccccc}
\toprule
\multirow{2}{*}{Method}  & \multicolumn{4}{l}{  \makebox[0.22\textwidth][c]{NC4K (4,121)} } & \multicolumn{4}{l}{\makebox[0.22\textwidth][c]{COD10K-Test (2,026)}} & \multicolumn{4}{l}{ \makebox[0.22\textwidth][c]{CAMO-Test (250)}}  \\ \cmidrule (lr){2-5}
\cmidrule (lr){6-9}\cmidrule (lr){10-13}
& \makecell{S$_m$ $\uparrow$} &\makecell{$\alpha $E $\uparrow$}  &\makecell{$w$F  $\uparrow$} &\makecell{M$\downarrow$} & \makecell{S$_m$  $\uparrow$} &\makecell{$\alpha $E  $\uparrow$}  &\makecell{$w$F  $\uparrow$} &\makecell{ M$\downarrow$} & \makecell{S$_m$  $\uparrow$} &\makecell{$\alpha$E  $\uparrow$}  &\makecell{$w$F  $\uparrow$} &\makecell{M$\downarrow$} \\ \midrule
\multicolumn{13}{l}{\emph{CNN-Based Methods}}\\ \midrule
$ \rm \textbf{PraNet}_{2020}$~\cite{fan2020pranet}&0.822 &0.871&0.724&0.059&0.789&0.839&0.629&0.045& 0.769&0.833&0.663 &0.094      \\
 $\rm \textbf{SINet}_{2020}$~\cite{fan2020camouflaged}&   0.808&0.883 &0.723&0.058&0.776&0.867&0.631&0.043&0.745&0.825&0.644& 0.092      \\
$\rm \textbf{SLSR}_{2021}$~\cite{lv2021simultaneously}&  0.840&0.902&0.766&0.048&0.804&0.882&0.673&0.037&0.787&0.855& 0.696& 0.080  \\
$\rm \textbf{MGL-R}_{2021}$~\cite{zhai2021mutual}&0.833&0.893&0.739&0.053&0.814&0.865&0.666& 0.035& 0.782&0.847&0.695&0.085 \\
$\rm \textbf{PFNet}_{2021}$~\cite{mei2021camouflaged} &0.829&0.892&0.745&0.053&0.800&0.868  &0.660&0.040&0.782&0.852&0.695&0.085       \\
$\rm \textbf{UJSC}_{2021}$~\cite{li2021uncertainty}&0.842&0.907&0.771&0.047&0.809&0.891&0.684&0.035& 0.800&0.853&0.728 &0.073\\
$\rm \textbf{C}^2\textbf{FNet}_{2021}$~\cite{suncontext}&  0.838&0.898&0.762& 0.049&0.813 &0.886  &0.686& 0.036& 0.796&0.864&0.719&0.080\\
$\rm \textbf{SINetV2}_{2022}$~\cite{fan2022concealed}& 0.847 & 0.898& 0.770 & 
0.048 & 0.815 & 0.863  &0.680 & 0.037  & 0.820 &0.875 &0.743    &0.070   \\
$\rm \textbf{DGNet}_{2022}$~\cite{ji2022deep}& 0.857& 0.907 & 0.784 & 0.042 & 0.822 &0.877  & 0.693 & 0.033  &0.839 &0.901 &0.769    &0.057   \\
$\rm \textbf{SegMaR}_{2022}$~\cite{Jia_2022_CVPR}& 0.841&0.905& 0.781&0.046& 0.833 &0.895  &0.724 & 0.033  &0.815 &0.872 &0.742    &0.071   \\
$\rm \textbf{ZoomNet}_{2022}$~\cite{Pang_2022_CVPR}&0.853&0.907&0.784&0.043&0.838&0.893&0.729&0.029&0.820&0.883&0.752& 0.066       \\
$\rm \textbf{FDNet}_{2022}$~\cite{Zhong_2022_CVPR}&0.834&0.895& 0.750&0.052& 0.837 & 0.897&0.731 & 0.030 &0.844&0.903&0.778&0.062   \\
\textbf{\nameofmethod{}-R (Ours)}&0.857&0.915&0.793&0.041&0.838&0.898&0.730&0.029&0.817&0.884&0.756&0.066 \\ 
\textbf{\nameofmethod{}-C (Ours)}&0.884&0.936&0.833&0.033&0.860&0.923&0.767&0.024&0.860&0.920&0.811&0.051 \\ \midrule
\multicolumn{13}{l}{\emph{Transformer-Based Methods}}\\ \midrule
$\mathrm{\textbf{COS-T}_{2021}}$~\cite{wang2021camouflaged}&0.825&0.881&0.730&0.055&0.790&0.901 &0.693&0.035&0.813&0.896&0.776&0.060\\
$\mathrm{\textbf{VST}_{2021}}$~\cite{liu2021visual}&0.830&0.887&0.740&0.053&0.810&0.866&0.680&0.035&0.805&0.863&0.780&0.069\\
$\rm \textbf{UGTR}_{2021}$~\cite{yang2021uncertainty}& 0.839&0.886& 0.746& 0.052&  0.817& 0.850 &0.666 & 0.036  & 0.784& 0.859& 0.794&0.086       \\
$\mathrm{\textbf{ICON}_{2022}}$~\cite{zhuge2022salient}&0.858&0.914&0.782&0.041&0.818&0.882&0.688&0.033&0.840&0.902&0.769&0.058\\
$\mathrm{\textbf{TPRNet}_{2022}}$~\cite{zhang2022tprnet}&0.854&0.903&0.790&0.047&0.829&0.892&0.725&0.034&0.814&0.870&0.781&0.076\\
$\mathrm{\textbf{DTINet}_{2022}}$~\cite{liu2022boosting}&0.863&0.915&0.792&0.041&0.824&0.893&0.695&0.034&0.857&0.912&0.796&0.050 \\
\textbf{\nameofmethod{}-S 
 (Ours)}&0.888&\highlight{0.941}&0.840&0.031&0.862&\highlight{0.932}&0.772&0.024&\highlight{0.876}&\highlight{0.935}&\highlight{0.832}&\highlight{0.043}\\ 
\textbf{\nameofmethod{}-P (Ours)}&  \highlight{0.892} & \highlight{0.941}& \highlight{0.847}&\highlight{0.030}&\highlight{0.869}&0.931&\highlight{0.786}&\highlight{0.023}&0.872&0.931&0.831&0.046 \\ 
\bottomrule %
\end{tabular}
\vspace{-5pt}
\caption{
Comparison of our \nameofmethod{} with the recent \sArt methods. 
 `-R': ResNet~\cite{he2016deep}, 
 `-C': ConvNext~\cite{liu2022convnet}, `-S': Swin Transformer~\cite{liu2021swin}, `-P': PVTv2~\cite{wang2022pvt}. As can be seen, our \nameofmethod{}-P performs much better than previous methods with either CNN- or Transformer-based models.
 `$\uparrow$': the higher the better, `$\downarrow$': the lower the better. 
}
\vspace{-5pt}
\label{tab:main_performance}
\end{table*}

\begin{figure}[t!]
  \centering
  \includegraphics[width=0.95\linewidth]{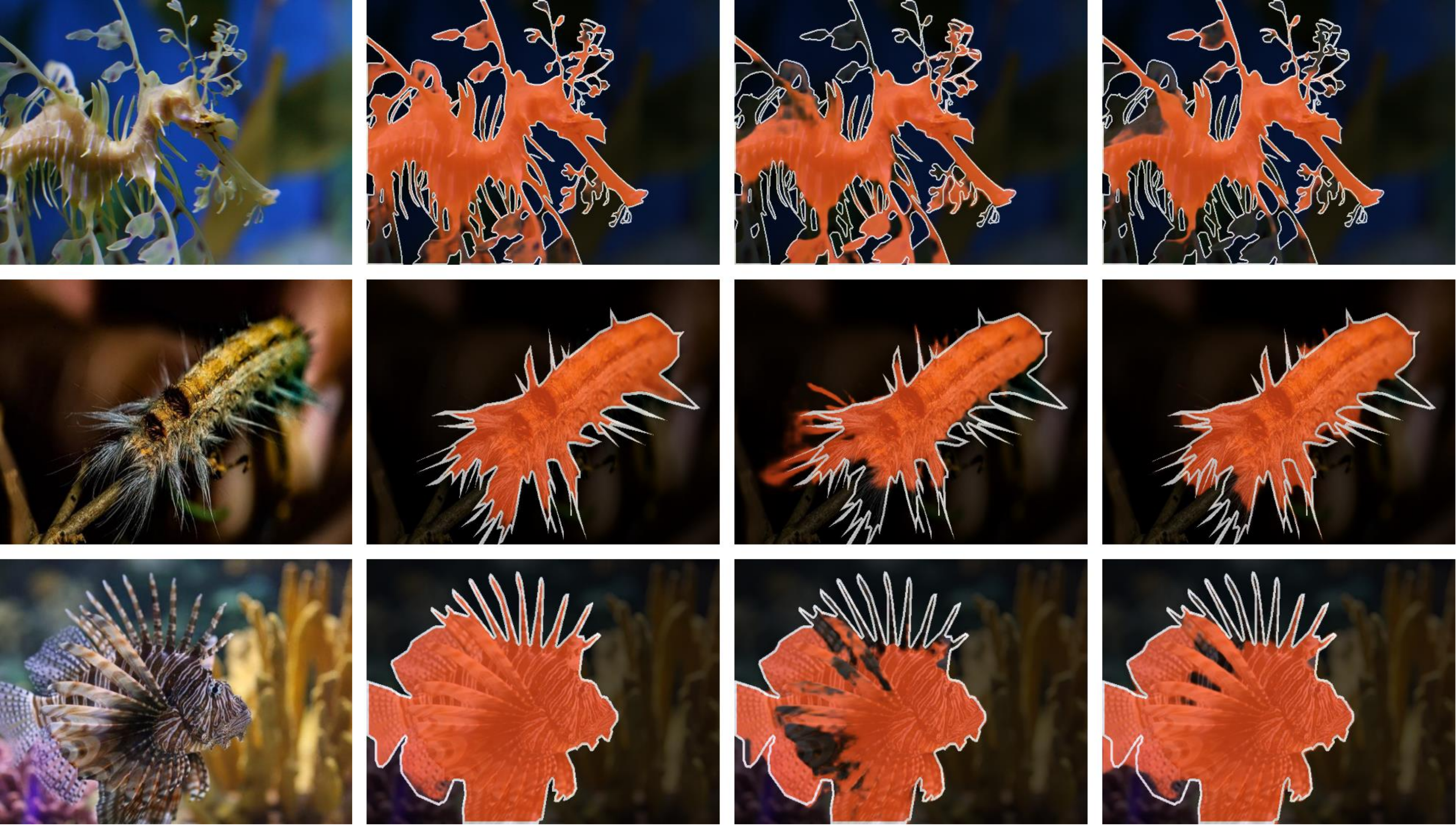}
  \put (-209.5, -10){\footnotesize{Image}}
  \put (-150.5, -10){\footnotesize{Ours}}
  \put (-105.5, -10){\footnotesize{SegMaR~\cite{Jia_2022_CVPR}}}
  \put (-49.5, -10){\footnotesize{ZoomNet~\cite{Pang_2022_CVPR}}}
  \vspace{-5pt}
  \caption{Comparisons of our \nameofmethod{} and other SOTA methods on the borders of segmentation.
  The borders of GT are marked in white, and the ones of predictions are in orange. }\label{fig:borders}
  \vspace{-5pt}
\end{figure}
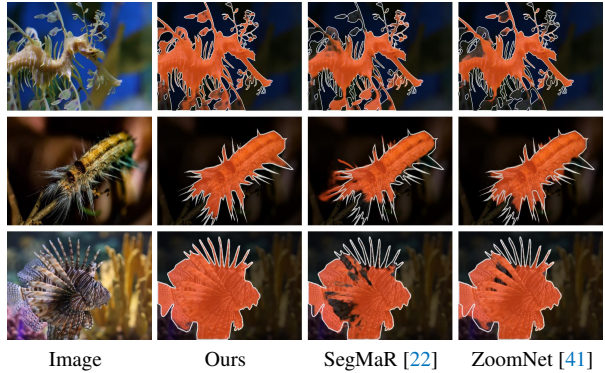

\subsection{Qualitative Evaluation}
\myPara{Visualization of Predictions.}
\figref{fig:vis_method} presents the visualization samples of our \nameofmethod{} and three previous \sArt methods. 
In order to better show the performance of these models, several typical samples containing different complex scenarios in the COD field are selected.
As shown in the top row, other methods are struggled to precisely perceive the camouflaged objects from their similar surroundings.
Sometimes, it is also difficult for them to identify the camouflaged objects owing to the lack of global contrast information, as shown in the bottom two rows.
In short, they misjudge some regions or miss parts of the targets when dealing with complex conditions due to the lack of a comprehensive understanding of the foreground and background.
%
In contrast, by explicitly perceiving the foreground-background clues, our \nameofmethod{} can generate high-quality segmentation maps of the camouflaged objects even under difficult conditions.

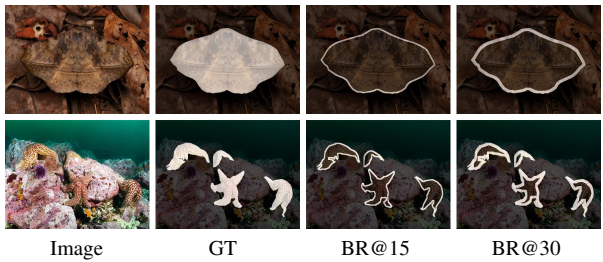
\begin{figure}[tp]
  \centering
  \includegraphics[width=0.95\linewidth]{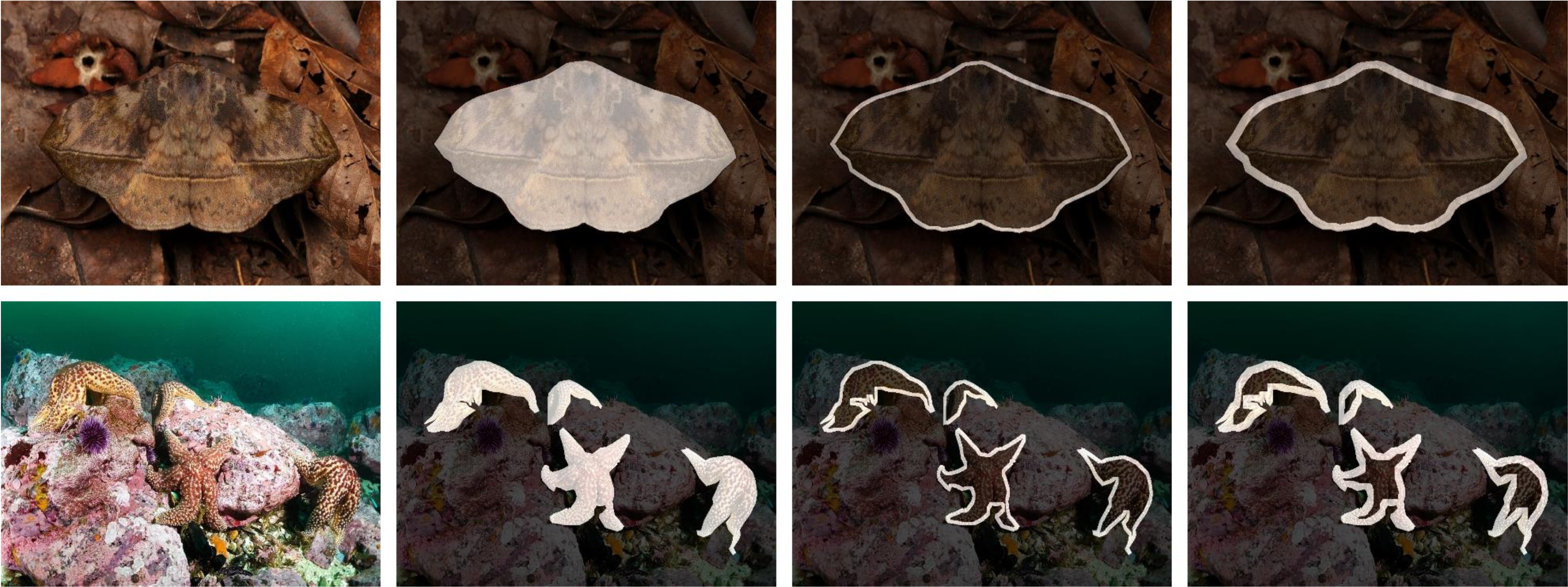}
  \put (-208, -10){\footnotesize{Image}}
  \put (-148, -10){\footnotesize{GT}}
  \put (-98, -10){\footnotesize{BR@15}}
  \put (-41, -10){\footnotesize{BR@30}}
  \vspace{-5pt}
  \caption{Illustration for the region of borders.
    `BR@15' and `BR@30' are border regions that are generated by dilating the borders of GT via $15\times 15$ and $30\times 30$ dilation kernels respectively.}
  \label{fig:borders-metric}
\end{figure}

\begin{table}[tp]
    \small
    \setlength\tabcolsep{2pt}
    \renewcommand{\arraystretch}{0.9}
    \begin{tabular}{lcccc} \toprule
    \multirow{2}{*}{Method} & \multicolumn{2}{c}{\makecell{BR@15}} & \multicolumn{2}{c}{\makecell{BR@30}} \\ \cmidrule (lr){2-3} \cmidrule (lr){4-5}

    &\makecell{BR-$w$F $\uparrow$} &\makecell{BR-M$\downarrow$}   &\makecell{BR-$w$F $\uparrow$} &\makecell{BR-M$\downarrow$}   \\ \midrule
    $\textbf{SINet}_{2020}$~\cite{fan2020camouflaged} & \makecell{0.645}& \makecell{0.103} &\makecell{0.707}    & \makecell{0.069}   \\
    $\textbf{PFNet}_{2021}$~\cite{mei2021camouflaged} & \makecell{0.677}&\makecell{0.091}  & \makecell{0.736}     & \makecell{0.064}   \\ 
    $\textbf{ZoomNet}_{2022}$~\cite{Pang_2022_CVPR} & \makecell{0.716}& \makecell{0.074} &\makecell{0.762}      & \makecell{0.059}   \\ 
    $\textbf{UGTR}_{2021}$~\cite{yang2021uncertainty} &0.672&0.084&0.720&0.063  \\ 
    $\textbf{DTINet}_{2022}$~\cite{liu2022boosting}&0.726&0.074&0.770&0.054\\
    \textbf{\nameofmethod{}-P (Ours)}&\makecell{\highlight{0.777}}&\makecell{\highlight{0.061}}&\makecell{\highlight{0.818}}&\makecell{\highlight{0.043}}
    \\\bottomrule  
    \end{tabular}
    \vspace{-5pt}
    \caption{Comparison of our \nameofmethod{}-P with other methods in terms of BR-$w\mathrm{F}$ and BR-M on the NC4K dataset~\cite{lv2021simultaneously}.}
    \vspace{-8pt}
    \label{tab:border}
\end{table}

\myPara{Object Border Quality Comparison.}
Camouflaged objects sometimes possess peculiar-looking shapes, as shown
in~\figref{fig:borders}.
To demonstrate how well our \nameofmethod{} performs when coping with
these kinds of objects, we show some prediction results in~\figref{fig:borders} and depict the GT object borders
with white curves.
The borders of our predictions are closer to those of the GT objects, while there are obvious deviations in the predictions by other methods. 
These visualizations indicate that our model can segment more precise camouflaged targets.



\begin{table*}[!ht]
\setlength\tabcolsep{6pt}
\renewcommand{\arraystretch}{0.9}
\small
\centering
\setlength{\belowcaptionskip}{0cm}   
\resizebox{\linewidth}{!}{
\begin{tabular}{lllllllllllll}
\toprule
\multirow{2}{*}{Setting} & \multicolumn{4}{l}{  \makebox[0.22\textwidth][c]{NC4K (4,121)} } & \multicolumn{4}{l}{\makebox[0.22\textwidth][c]{COD10K-Test (2,026)}} & \multicolumn{4}{l}{ \makebox[0.22\textwidth][c]{CAMO-Test (250)}} \\  \cmidrule (lr){2-5}
\cmidrule (lr){6-9}\cmidrule (lr){10-13}
& \makecell{S$_m$ $\uparrow$} &\makecell{$\alpha $E $\uparrow$}  &\makecell{$w$F  $\uparrow$} &\makecell{M$\downarrow$} & \makecell{S$_m$  $\uparrow$} &\makecell{$\alpha $E  $\uparrow$}  &\makecell{$w$F  $\uparrow$} &\makecell{ M$\downarrow$} & \makecell{S$_m$  $\uparrow$} &\makecell{$\alpha$E  $\uparrow$}  &\makecell{$w$F  $\uparrow$} &\makecell{M$\downarrow$}  \\ \midrule
Baseline&0.859&0.916&0.801&0.043 &0.830& 0.904 &0.719 &0.032 &0.838& 0.891 &0.780 &0.058  \\
Baseline+MSA& 0.875&	0.926	&0.808&	0.036	&0.848&	0.906&	0.735&	0.029	&0.858	&0.918&	0.798&	0.052
\\ \midrule
\textbf{\nameofmethod{}-P} w/ TA only&0.880&	0.925	&0.825&	0.034&	0.850&	0.908&	0.749&	0.026	&0.862	&0.918	&0.812	&0.055
\\
\textbf{\nameofmethod{}-P}&  \highlight{0.892} & \highlight{0.941}& \highlight{0.847}&\highlight{0.030}&\highlight{0.869}&\highlight{0.931}&\highlight{0.786}&\highlight{0.023}&\highlight{0.872}&\highlight{0.931}&\highlight{0.831}&\highlight{0.046} \\
 \bottomrule
\end{tabular}}
\vspace{-8pt}
\caption{Ablation study of our \nameofmethod{} variants.
`Baseline': the transformer backbone and several convolution layers;
`+MSA': Baseline with MSA;
`w/TA only': Baseline equipped with TA and iterative refinement fashion.
}
\label{tab:ab1}
\end{table*}

\begin{table*}[bhtp]
    \setlength\tabcolsep{6pt}
    \renewcommand{\arraystretch}{0.9}
    \small
    \centering
    \setlength{\belowcaptionskip}{0cm}   
    \resizebox{\linewidth}{!}{
    \begin{tabular}{ccccllllllllllll}
    \toprule
    & \multicolumn{3}{c}{Decoder} &\multicolumn{4}{l}{  \makebox[0.22\textwidth][c]{NC4K (4,121)} } & \multicolumn{4}{l}{\makebox[0.22\textwidth][c]{COD10K-Test (2,026)}} & \multicolumn{4}{l}{ \makebox[0.22\textwidth][c]{CAMO-Test (250)}} \\  \cmidrule (lr){2-4}
    \cmidrule (lr){5-8}\cmidrule (lr){9-12}\cmidrule (lr){13-16}
    Settings & TA & F-TA & B-TA & \makecell{S$_m$ $\uparrow$} &\makecell{$\alpha $E $\uparrow$}  &\makecell{$w$F  $\uparrow$} &\makecell{M$\downarrow$} & \makecell{S$_m$  $\uparrow$} &\makecell{$\alpha $E  $\uparrow$}  &\makecell{$w$F  $\uparrow$} &\makecell{ M$\downarrow$} & \makecell{S$_m$  $\uparrow$} &\makecell{$\alpha$E  $\uparrow$}  &\makecell{$w$F  $\uparrow$} &\makecell{M$\downarrow$}  \\ \midrule
    1 & & &  &0.865& 0.921& 0.795& 0.041& 0.836& 0.915& 0.724 &0.030 &0.857 &0.915 &0.797 &0.053    \\
    2 & \cmark & & & 0.880&	0.925	&0.825&	0.034&	0.850&	0.908&	0.749&	0.026	&0.862	&0.918	&0.812	&0.055\\
    3 & & \cmark &  &0.878	&0.924&	0.812&	0.036&	0.849&	0.909	&0.742&	0.028&	0.861	&0.916	&0.811	&0.054

\\
    4 & & & \cmark  &0.875&	0.920&	0.815&	0.038	&0.845	&0.915&	0.742&	0.029	&0.856	&0.914&	0.809	&0.055\\
    5 &  & \cmark & \cmark &0.880&	0.928&	0.827&	0.035&	0.847&	0.919&	0.754&	0.024&	0.862&	0.921&	0.812&	0.052
\\
    6 & \cmark &  & \cmark &0.885&0.933&0.839&0.033&0.860&	0.924	&0.763&	0.025	&0.866&	0.926&	0.827&	0.048
\\
    7 & \cmark & \cmark && 0.889	&0.939&	0.845&0.031&	0.866	&0.927	&0.773&	0.024&	0.871&	0.929&	\highlight{0.832}&	\highlight{0.046}
\\
    8 & \cmark& \cmark & \cmark &  \highlight{0.892} & \highlight{0.941}& \highlight{0.847}&\highlight{0.030}&\highlight{0.869}&\highlight{0.931}&\highlight{0.786}&\highlight{0.023}&\highlight{0.872}&\highlight{0.931}&0.831&\highlight{0.046} \\ \bottomrule
    \end{tabular}}
    \vspace{-8pt}
    \caption{Ablation study on the proposed MSA. All three branches (`TA', `F-TA', and `B-TA') contribute to the overall performance. In addition, eliminating either the `F-TA' or `B-TA' branch hurts the performance.
    }
    \vspace{-5pt}
    \label{tab:ab2}
\end{table*}

\subsection{Quantitative Evaluation}\label{subsec:quan}
We compare our \nameofmethod{} with 12 CNN-based SOTA COD models, including ZoomNet~\cite{Pang_2022_CVPR}, FDNet~\cite{Zhong_2022_CVPR}, SegMaR~\cite{Jia_2022_CVPR}, DGNet~\cite{ji2022deep}, SINetV2~\cite{fan2022concealed}, $\rm C^2 FNet$~\cite{suncontext}, UJSC~\cite{li2021uncertainty}, PFNet~\cite{mei2021camouflaged}, MGL-R~\cite{zhai2021mutual}, SLSR~\cite{lv2021simultaneously}, SINet~\cite{fan2020camouflaged}, and PraNet~\cite{fan2020pranet} and 6 Transformer-based methods, including COS-T~\cite{wang2021camouflaged},  TPRNet~\cite{zhang2022tprnet}, VST~\cite{liu2021visual}, DTINet~\cite{liu2022boosting}, UGTR~\cite{yang2021uncertainty}, and ICON~\cite{zhuge2022salient}.
For a fair comparison, the prediction results are directly provided by their authors or generated by their well-trained models.
%

\myPara{Performance on Object Regions.}
As shown in~\tabref{tab:main_performance}, our proposed \nameofmethod{} consistently and significantly surpasses the previous methods on all three benchmarks without any post-process tricks or extra data for training. 
Compared to the recent CNN-based COD methods, such as ZoomNet~\cite{Pang_2022_CVPR}, FDNet~\cite{Zhong_2022_CVPR}, and SegMaR~\cite{Jia_2022_CVPR}, although they adopt strategies, like multi-stage training and inference that cost extra computational burden, our \nameofmethod{} still outperforms them on all benchmarks by a large margin. 
Meanwhile, compared to the Transformer-based models (\eg, TPRNet~\cite{zhang2022tprnet} and DTINet~\cite{liu2022boosting}), our method also performs better than them, setting new \sArt records.

\myPara{Performance on Border Regions.}
The borders of camouflaged targets are challenging to detect due to their irregular shapes and high similarity with their surroundings. 
In order to quantify the segmentation performance near the border regions, we calculate the $w\mathrm{F}$ and MAE scores in the border regions, denoted as BR-$w\mathrm{F}$ and BR-M, respectively. 
We attain the border regions by dilating the boundaries of the GT objects, as shown in~\figref{fig:borders-metric}. 
Note that the area of the region depends on the kernel size of the dilation operation. 
\tabref{tab:border} shows the performance calculated in the border regions `BR@15' and `BR@30'. 
Remarkably, \nameofmethod{} achieves much better results than other methods, demonstrating that our predictions perform better at GT object boundaries, while the predictions of other methods are significantly biased.


\subsection{Model Sweep}\label{subsec:ablation}
\myPara{Overall Results.}
We first ablate the network architecture of \nameofmethod{}.
The results are shown in \tabref{tab:ab1}.
The `Baseline' refers to the model with only the encoder followed by a convolution
for prediction.
When our MSA is applied, the performance can be clearly improved in terms of all evaluation metrics compared to the `Baseline'.
Then, we attempt to add the decoder with only TA left.
This progressive fusion strategy also helps compared to the `Baseline'.
Finally, we add our MSA in the decoder as in~\figref{fig:network}.
We can see that the performance can be further improved.
Both the top and bottom halves indicate the importance of MSA for COD.

\begin{figure*}[htp]
\setlength{\abovecaptionskip}{0.6cm}
  \centering
  \includegraphics[width=0.96\linewidth]{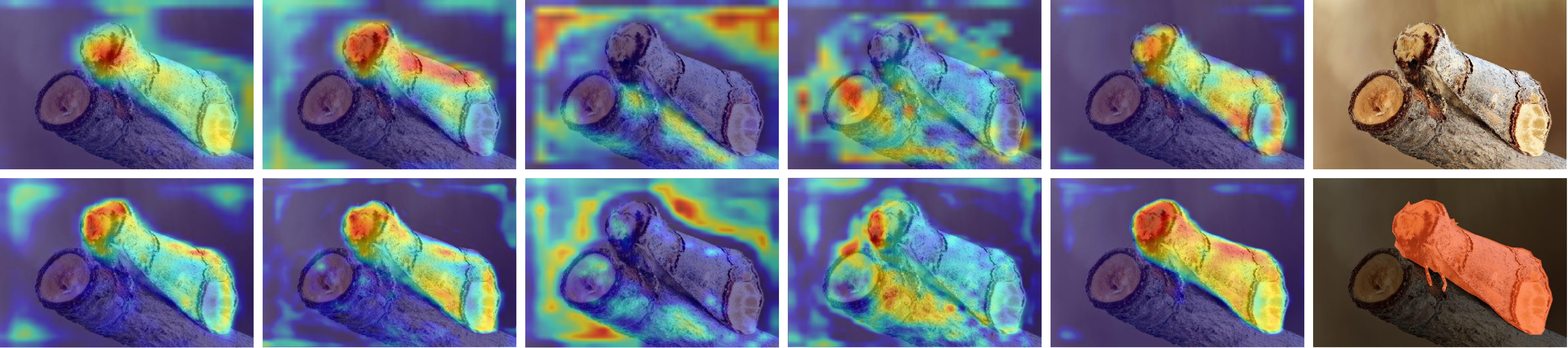}
  \put (-474,58){{\color{white}{\footnotesize{Input}}}}
  \put (-474,3){{\color{white}{\footnotesize{Input}}}}
  \put (-394,58){{\color{white}{\footnotesize{F-TA}}}}
  \put (-394,3){{\color{white}{\footnotesize{F-TA}}}}
  \put (-314,58){{\color{white}{\footnotesize{B-TA}}}}
  \put (-314,3){{\color{white}{\footnotesize{B-TA}}}}
  \put (-234,58){{\color{white}{\footnotesize{TA}}}}
  \put (-234,3){{\color{white}{\footnotesize{TA}}}}
  \put (-154,58){{\color{white}{\footnotesize{Output ($D_5$)}}}}
  \put (-154,3){{\color{white}{\footnotesize{Output ($D_2$)}}}}
  \put (-74,58){{\color{white}{\footnotesize{Image}}}}
  \put (-74,3){{\color{white}{\footnotesize{GT}}}}
  \put (-486,68){\rotatebox{90}{\footnotesize{\textbf{Stage-5}}}}
  \put (-486,14){\rotatebox{90}{\footnotesize{\textbf{Stage-2}}}}
  \vspace{-12pt}
  \caption{Visualization of the feature maps around MSA. 
  The features from Stage 2 and Stage 5 are chosen for comparison. 
  }\label{fig:feature_vis}
\end{figure*}

\begin{figure*}[htp]
  \centering
  \includegraphics[width=0.95\linewidth]{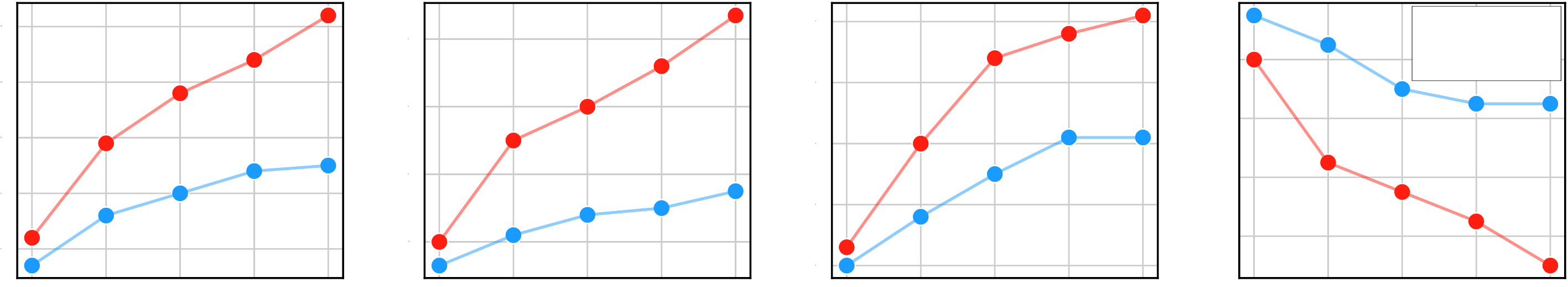}
  \put (-422, 88){\small{S$_m$}}
  \put (-479, 75.6){\footnotesize{.89}}
  \put (-479, 59){\footnotesize{.88}}
  \put (-479, 43){\footnotesize{.87}}
  \put (-479, 26){\footnotesize{.86}}
  \put (-479, 9.4){\footnotesize{.85}}
  \put (-462, -7){\footnotesize{5}}
  \put (-440, -7){\footnotesize{4}}
  \put (-417, -7){\footnotesize{3}}
  \put (-395, -7){\footnotesize{2}}
  \put (-373, -7){\footnotesize{1}}
  \put (-177.5, 88){\small{$\alpha \mathrm{E}$}}
  \put (-234, 77.2){\footnotesize{.94}}
  \put (-234, 59.2){\footnotesize{.93}}
  \put (-234, 40.5){\footnotesize{.92}}
  \put (-234, 22.3){\footnotesize{.91}}
  \put (-234, 3.8){\footnotesize{.90}}
  \put (-218, -7){\footnotesize{5}}
  \put (-196, -7){\footnotesize{4}}
  \put (-173, -7){\footnotesize{3}}
  \put (-151, -7){\footnotesize{2}}
  \put (-129, -7){\footnotesize{1}}
  \put (-302, 88){\small{$w\mathrm{F}$}}
  \put (-356, 72){\footnotesize{.84}}
  \put (-356, 52){\footnotesize{.82}}
  \put (-356, 31.8){\footnotesize{.80}}
  \put (-356, 11.6){\footnotesize{.78}}
  \put (-340, -7){\footnotesize{5}}
  \put (-318, -7){\footnotesize{4}}
  \put (-295, -7){\footnotesize{3}}
  \put (-273, -7){\footnotesize{2}}
  \put (-251, -7){\footnotesize{1}}
  \put (-55, 88){\small{M}}
  \put (-115, 66){\footnotesize{.044}}
  \put (-115, 48.2){\footnotesize{.040}}
  \put (-115, 30.4){\footnotesize{.036}}
  \put (-115, 12.6){\footnotesize{.032}}
  \put (-96, -7){\footnotesize{5}}
  \put (-74, -7){\footnotesize{4}}
  \put (-51, -7){\footnotesize{3}}
  \put (-29, -7){\footnotesize{2}}
  \put (-7, -7){\footnotesize{1}}
  \put (-448, 47){\footnotesize{.869}}
  \put (-448, 25){\footnotesize{.856}}
  \put (-384, 72){\footnotesize{.892}}
  \put (-384, 40){\footnotesize{.865}}
  \put (-261.5, 72){\footnotesize{.847}}
  \put (-261.5, 30){\footnotesize{.795}}
  \put (-326, 48){\footnotesize{.810}}
  \put (-326, 18){\footnotesize{.782}}
  \put (-202, 48){\footnotesize{.920}}
  \put (-202, 25){\footnotesize{.908}}
  \put (-138, 73){\footnotesize{.941}}
  \put (-138, 48){\footnotesize{.921}}
  \put (-79, 76){\footnotesize{.045}}
  \put (-79, 41){\footnotesize{.037}}
  \put (-16, 46){\footnotesize{.041}}
  \put (-16, 10){\footnotesize{.030}}
  \put (-44,77){\footnotesize{{\color{red}{\textbf{\large{--}}}} w/ MSA}}
  \put (-44,65){\footnotesize{{\color{cyan}{\textbf{\large{--}}}} w/o MSA}}
  \caption{Performance on the NC4K dataset at different feature levels of our progressive fusion strategy. The proposed MSA fits well with the decoder in that the performance gap tends to be larger from feature level 5 to feature level 1.
  }\label{fig:stages}
\end{figure*}
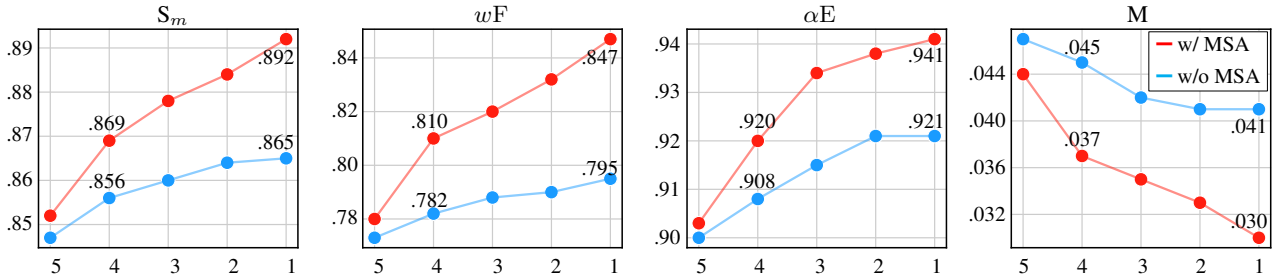

\myPara{Masked Separable Attention.}
We then ablate how each component in MSA helps.
\tabref{tab:ab2} shows the experimental results. 
The `Baseline' in the first row can be viewed as a simple feature pyramid network, \emph{i.e.,}  no MSA is added in \figref{fig:network}.
We can observe that each type of attention component is helpful to improve
performance.
%
Though TA yields more performance gain compared to F-TA and B-TA, combining
either F-TA or B-TA with TA can further improve the results.
In particular, adding all three components yields the best results on all three datasets.
This series of experiments indicate that separately processing the foreground and background with the proposed MSA is useful for segmenting the camouflaged objects.

\myPara{Feature Visualization.}
To provide more promising insights into our MSA, we also visualize the features around it. 
Note that the MSA of Stages 2 and 5 are chosen for visualization. 
As shown in~\figref{fig:feature_vis}, F-TA and B-TA are able to effectively obtain the cues of the foreground and background. 
The features from F-TA, B-TA, and TA are complementary to form a completely camouflaged object.
Consequently, the precise location and detailed information of the camouflaged object can be captured easily. 
In addition, compared to the features in Stage 5, we can distinguish the camouflaged targets more clearly according to the ones in Stage 2, which indicates that the progressive refinement decoder can fully exploit the potential of our MSA.

\myPara{MSA in Progressive Fusion.}
In our model, a progressive fusion strategy is adopted to gradually refine
the segmentation map in the decoder.
To show the impact of our MSA in this strategy, we depict the performance curves at different feature levels under the setup of w/ MSA and w/o MSA.
The results can be found in \figref{fig:stages}.
%
%
We can see that from feature level 5 to feature level 1, the performance gap between
the model without MSA (blue line) and the one with MSA tends to be more significant for all four
evaluation metrics.
This indicates that the proposed MSA is compatible with the progressive fusion decoder.

%

\begin{table}[t!]
\setlength\tabcolsep{5pt}
\small
\centering
\setlength{\belowcaptionskip}{0cm}   
\renewcommand{\arraystretch}{1}
\begin{tabular}{lcccccc}
\toprule
\multirow{2}{*}{Settings} & \multicolumn{2}{l}{\makecell{Computations}} & \multicolumn{2}{l}{\makecell{NC4K } }&\multicolumn{2}{l}{\makecell{COD10K-Test }}\\  \cmidrule (lr){2-3}  \cmidrule (lr){4-5}  \cmidrule (lr){6-7}
& \makecell{Params}& \makecell{MAC}&  \makecell{S$_m$ $\uparrow$}&\makecell{$w$F $\uparrow$}  &  \makecell{S$_m$ $\uparrow$}  &\makecell{$w$F $\uparrow$} \\ \midrule
 $C_d=32$&63M&30G&0.887&0.842&0.865&0.779\\
 $C_d=64$&65M&34G&0.890&0.846&\highlight{0.869}&0.784\\
 $C_d=128$ &71M&47G&\highlight{0.892}&0.847&\highlight{0.869} &\highlight{0.786}\\
 $C_d=192$&82M&69G&\highlight{0.892}&\highlight{0.849}&0.868&\highlight{0.786}\\
 $C_d=256$ &97M&99G& 0.891&0.846&0.866&0.784\\ \bottomrule
\end{tabular}
\vspace{-8pt}
\caption{Ablation study on the channel numbers in the decoder. 
All the variants are equipped with our MSA. 
More comparisons are shown in the supplementary materials.
}
\vspace{-5pt}
\label{tab:ab_channel}
\end{table}

\myPara{Decoder Width.}
The width (\#channels) of the decoder affects not only the model size but
also the inference speed.
\tabref{tab:ab_channel} shows the changes in model parameters, computational cost, and performance when the number of channels changes.
We can see a clear improvement in our model when $C_d$ increases from 32 to 128. 
However, when the width changes from 128 to 192, the performance improves little, but the parameters and computations rise.
As a result, $C_d$ is set to 128 for the trade-off between efficiency and model performance. 
%
%
%
%
This is different from most previous FPN-based methods~\cite{hou2019deeply,liu2019simple} that use larger channel numbers at deeper feature levels.

\section{Conclusions}
We present \nameofmethod{} for camouflaged object segmentation. 
The core of our \nameofmethod{} is the masked separable attention (MSA) that
separately deals with the foreground and background regions using different attention heads.
%
%
To make better use of our MSA, we adopt a progressive refinement decoder to gradually improve the segmentation quality at different feature levels in a top-down manner.
Extensive experiments show that \nameofmethod{} surpasses the existing 18 \sArt models with clear improvements.
%
We hope our MSA can facilitate the development of COD and other binary segmentation tasks.

\begin{figure*}[htp!]
\setlength{\abovecaptionskip}{0.4cm}
  \centering
  \includegraphics[width=0.99\linewidth]{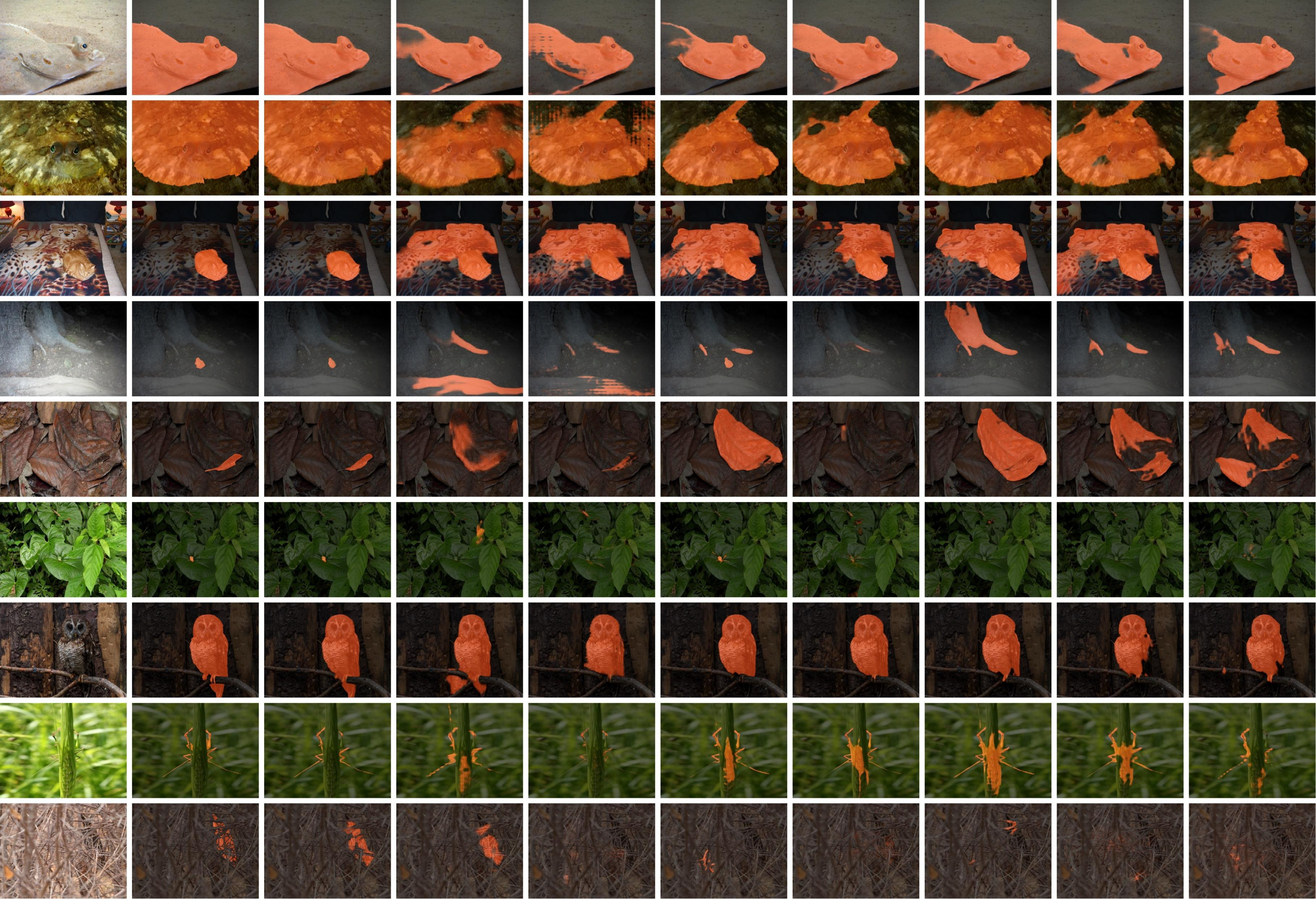}
  \vspace{-5pt}
  \put(-480,-10){\footnotesize{Image}}
  \put(-428,-10){\footnotesize{GT}}
  \put(-380,-10){\footnotesize{Ours}}
  \put(-341.4,-10){\footnotesize{DTINet}~\cite{liu2022boosting}}
  \put(-290,-10){\footnotesize{UGTR}~\cite{yang2021uncertainty}}
  \put(-243,-10){\footnotesize{SegMaR}~\cite{Jia_2022_CVPR}}
  \put(-196,-10){\footnotesize{ZoomNet}~\cite{Pang_2022_CVPR}}
  \put(-141,-10){\footnotesize{UJSC}~\cite{li2021uncertainty}}
  \put(-92,-10){\footnotesize{PFNet}~\cite{mei2021camouflaged}}
  \put(-39.5,-10){\footnotesize{SINet}~\cite{fan2020camouflaged}}
  \caption{More visual example comparisons between our \nameofmethod{} and some previous SOTA methods. Segmentation results are highlighted in orange. 
  }\label{fig:supp_vis}
\end{figure*}
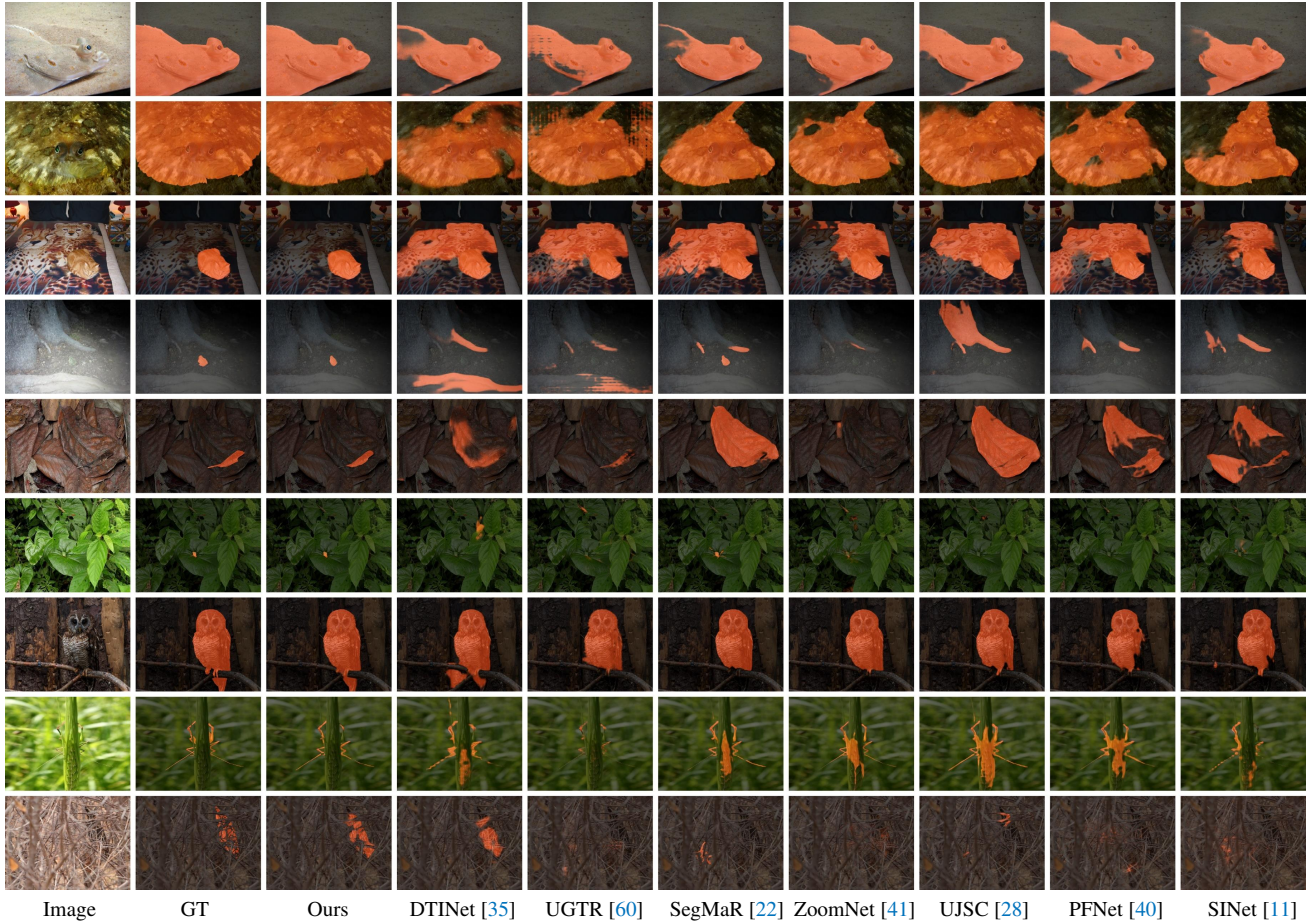

\begin{table*}[htp!]
\setlength\tabcolsep{5.6pt}
\small
\centering
\setlength{\belowcaptionskip}{0cm}   
\renewcommand{\arraystretch}{0.9}
\begin{tabular}{lcccccccccccccc}
\toprule
\multirow{2}{*}{Settings} & \multicolumn{2}{l}{\makebox[0.13\textwidth][c]{\makecell{Computations}}} & \multicolumn{4}{c}{\makecell{NC4K} }&\multicolumn{4}{c}{\makecell{COD10K-Test}}&\multicolumn{4}{c}{\makecell{CAMO}}\\  \cmidrule(lr){2-3}  \cmidrule(lr){4-7}  \cmidrule(lr){8-11} \cmidrule(lr){12-15}
& \makecell{Params}& \makecell{MAC}&  \makecell{S$_m$ $\uparrow$}&\makecell{$\alpha$E}&\makecell{$w$F $\uparrow$} &\makecell{M} &  \makecell{S$_m$ $\uparrow$}&\makecell{$\alpha$E}&\makecell{$w$F $\uparrow$} &\makecell{M} &  \makecell{S$_m$ $\uparrow$}&\makecell{$\alpha$E}&\makecell{$w$F $\uparrow$} &\makecell{M} \\ \midrule
 $C_d=32$&63M&30G&0.887&0.938&0.842&0.031&0.865&0.925&0.779&0.024&0.871&0.926&0.827&0.047\\
 $C_d=64$&65M&34G&0.890&0.940&0.846&0.031&\highlight{0.869}&0.929&0.784&\highlight{0.023}&0.872&0.931&0.829&0.047\\
 $C_d=128$ &71M&47G&  \highlight{0.892} & 0.941&0.847&\highlight{0.030}&\highlight{0.869}&0.931&\highlight{0.786}&\highlight{0.023}&0.872&0.931&0.831&\highlight{0.046} \\
 $C_d=192$&82M&69G&\highlight{0.892}&\highlight{0.942}&\highlight{0.849}&\highlight{0.030}&0.868&\highlight{0.932}&\highlight{0.786}&\highlight{0.023}&\highlight{0.873}&\highlight{0.933}&\highlight{0.834}&\highlight{0.046}\\
 $C_d=256$ &97M&99G& 0.891&\highlight{0.942}&0.846&\highlight{0.030}&0.866&\highlight{0.932}&0.784&\highlight{0.023}&0.869&0.931&0.830&\highlight{0.046}\\ \bottomrule
\end{tabular}
\vspace{-8pt}
\caption{Detailed ablation study on the channel numbers in the decoder of our~\nameofmethod{}. 
}
\label{tab:ab_channel_supp}
\end{table*}

\section*{Acknowledgements}

Qibin Hou was partially supported by the Natural Science Foundation of China (No. 62276145) and the Fundamental Research Funds for the Central Universities (No. 63223049).


\appendix


\begin{table*}[tp]
\setlength\tabcolsep{3.5pt}
\renewcommand{\arraystretch}{0.9}
\small
\setlength{\belowcaptionskip}{0cm}   
\begin{tabular}{lccccccccccccccccc}
\toprule
\multirow{2}{*}{Method}  & \multicolumn{4}{c}{  \makebox[0.14\textwidth][c]{NC4K (4,121)} } & \multicolumn{4}{c}{\makebox[0.14\textwidth][c]{COD10K-Test (2,026)}} & \multicolumn{4}{c}{ \makebox[0.14\textwidth][c]{CAMO-Test (250)}}  & \multicolumn{4}{c}{ \makebox[0.14\textwidth][c]{CHAMELEON (76)}} \\ \cmidrule (lr){2-5}
\cmidrule (lr){6-9}\cmidrule (lr){10-13} \cmidrule (lr){14-17}
& \makecell{S$_m$ $\uparrow$} &\makecell{$\alpha $E $\uparrow$}  &\makecell{$w$F  $\uparrow$} &\makecell{M$\downarrow$} & \makecell{S$_m$  $\uparrow$} &\makecell{$\alpha $E  $\uparrow$}  &\makecell{$w$F  $\uparrow$} &\makecell{ M$\downarrow$} & \makecell{S$_m$  $\uparrow$} &\makecell{$\alpha$E  $\uparrow$}  &\makecell{$w$F  $\uparrow$} &\makecell{M$\downarrow$} &\makecell{S$_m$  $\uparrow$} &\makecell{$\alpha$E  $\uparrow$}  &\makecell{$w$F  $\uparrow$} &\makecell{M$\downarrow$} \\ \midrule
\multicolumn{17}{l}{\emph{CNN-Based Methods}}\\ \midrule
$ \rm \textbf{PraNet}_{2020}$~\cite{fan2020pranet}&.822 &.871&.724&.059&.789&.839&.629&.045& .769&.833&.663 &.094 &.860&.898&.763&.044     \\
 $\rm \textbf{SINet}_{2020}$~\cite{fan2020camouflaged}&   .808&.883 &.723&.058&.776&.867&.631&.043&.745&.825&.644& .092 &.872&.938&.806&.034     \\
$\rm \textbf{SLSR}_{2021}$~\cite{lv2021simultaneously}&  .840&.902&.766&.048&.804&.882&.673&.037&.787&.855& .696& .080 &.890&.936&.822&.030 \\
$\rm \textbf{MGL-R}_{2021}$~\cite{zhai2021mutual}&.833&.893&.739&.053&.814&.865&.666& .035& .782&.847&.695&.085&.893&.923&.812&.031 \\
$\rm \textbf{PFNet}_{2021}$~\cite{mei2021camouflaged} &.829&.892&.745&.053&.800&.868  &.660&.040&.782&.852&.695&.085&.882&.942&.810&.033       \\
$\rm \textbf{UJSC}_{2021}$~\cite{li2021uncertainty}&.842&.907&.771&.047&.809&.891&.684&.035& .800&.853&.728 &.073&.891&.943&.833&.030\\
$\rm \textbf{C}^2\textbf{FNet}_{2021}$~\cite{suncontext}&  .838&.898&.762& .049&.813 &.886  &.686& .036& .796&.864&.719&.080&.888&.932&.828&.032\\
$\rm \textbf{SINetV2}_{2022}$~\cite{fan2022concealed}& .847 & .898& .770 & 
.048 & .815 & .863  &.680 & .037  & .820 &.875 &.743    &.070 &.888&.930&.816&.030  \\
$\rm \textbf{DGNet}_{2022}$~\cite{ji2022deep}& .857& .907 & .784 & .042 & .822 &.877  & .693 & .033  &.839 &.901 &.769    &.057 &.890&.934&.816&.029  \\
$\rm \textbf{SegMaR}_{2022}$~\cite{Jia_2022_CVPR}& .841&.905& .781&.046& .833 &.895  &.724 & .033  &.815 &.872 &.742    &.071  &.897&.950&.835&.027 \\
$\rm \textbf{ZoomNet}_{2022}$~\cite{Pang_2022_CVPR}&.853&.907&.784&.043&.838&.893&.729&.029&.820&.883&.752& .066 &.902&.952&.845&.023      \\
$\rm \textbf{FDNet}_{2022}$~\cite{Zhong_2022_CVPR}&.834&.895& .750&.052& .837 & .897&.731 & .030 &.844&.903&.778&.062 &.894&.948&.819&.030  \\
\textbf{\nameofmethod{}-R (Ours)}&.857&.915&.793&.041&.838&.898&.730&.029&.817&.884&.756&.066&.900&.949&.843&.024 \\ 
\textbf{\nameofmethod{}-C (Ours)}&.884&.936&.833&.033&.860&.923&.767&.024&.860&.920&.811&.051&.901&.955&.846&.025& \\ \midrule
\multicolumn{17}{l}{\emph{Transformer-Based Methods}}\\ \midrule
$\mathrm{\textbf{COS-T}_{2021}}$~\cite{wang2021camouflaged}&.825&.881&.730&.055&.790&.901 &.693&.035&.813&.896&.776&.060&.885&.948&.854&.025\\
$\mathrm{\textbf{VST}_{2021}}$~\cite{liu2021visual}&.830&.887&.740&.053&.810&.866&.680&.035&.805&.863&.780&.069&.888&.936&.820&.033\\
$\rm \textbf{UGTR}_{2021}$~\cite{yang2021uncertainty}& .839&.886& .746& .052&  .817& .850 &.666 & .036  & .784& .859& .794&.086      &.888&.921&.794&.031 \\
$\mathrm{\textbf{ICON}_{2022}}$~\cite{zhuge2022salient}&.858&.914&.782&.041&.818&.882&.688&.033&.840&.902&.769&.058&.854&.920&.763&.037\\
$\mathrm{\textbf{TPRNet}_{2022}}$~\cite{zhang2022tprnet}&.854&.903&.790&.047&.829&.892&.725&.034&.814&.870&.781&.076&.891&.930&.816&.031\\
$\mathrm{\textbf{DTINet}_{2022}}$~\cite{liu2022boosting}&.863&.915&.792&.041&.824&.893&.695&.034&.857&.912&.796&.050&.883&.928&.813&.033 \\
\textbf{\nameofmethod{}-S 
 (Ours)}&.888&\highlight{.941}&.840&.031&.862&\highlight{.932}&.772&.024&\highlight{.876}&\highlight{.935}&\highlight{.832}&\highlight{.043}&.891&.953&.829&.026\\ 
\textbf{\nameofmethod{}-P (Ours)}&  \highlight{.892} & \highlight{.941}& \highlight{.847}&\highlight{.030}&\highlight{.869}&.931&\highlight{.786}&\highlight{.023}&.872&.931&.831&.046 &\highlight{.910}&\highlight{.970}&\highlight{.865}&\highlight{.022}\\ 
\bottomrule %
\end{tabular}
\vspace{-5pt}
\caption{
Comparison of our \nameofmethod{} with the recent \sArt methods on four datasets. 
 `-R': ResNet~\cite{he2016deep}, 
 `-C': ConvNext~\cite{liu2022convnet}, `-S': Swin Transformer~\cite{liu2021swin}, `-P': PVTv2~\cite{wang2022pvt},
 `$\uparrow$': the higher the better, `$\downarrow$': the lower the better. 
}
\label{tab:supp_main_performance}
\end{table*}

\section{Performance on All Benchmark Datasets}

In the main paper, we have provided the evaluation results of our CamoFormer on the three most popular benchmark datasets, namely NC4K, COD10K-Test, and CAMO-Test. 
Here, we also report the results on CHAMELEON~\cite{skurowski2018animal}, which contains 76 image-annotation pairs.
As shown in ~\tabref{tab:supp_main_performance}, our \nameofmethod{} can improve all other methods significantly.

\section{More Visual Comparisons}
We show more visualization examples of our CamoFormer and the recent SOTA methods in~\figref{fig:supp_vis}. These examples cover a wide range of camouflaged scenes, such as big objects, small objects, occlusion, and indefinable boundaries. 
Obviously, our \nameofmethod{} performs much better than other methods.

\section{More Ablation Results of Decoder Channels}
The ablation results of the decoder channels on three benchmark datasets are shown in~\tabref{tab:ab_channel_supp}. 
We adopt $C_d=128$ for the trade-off between performance and computations. It is noteworthy that our CamoFormer can outperform other previous methods even if $C_d=32$ is applied. 

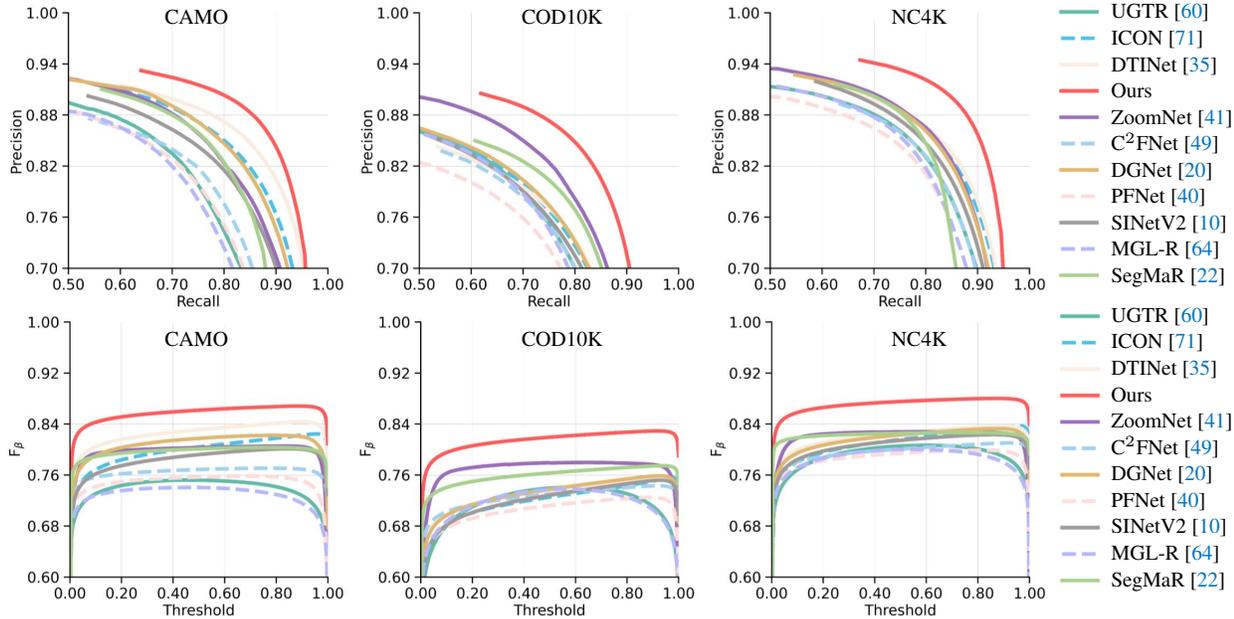
\begin{figure*}[htp!]
\setlength{\abovecaptionskip}{0.4cm}
  \centering
  \includegraphics[width=0.95\linewidth]{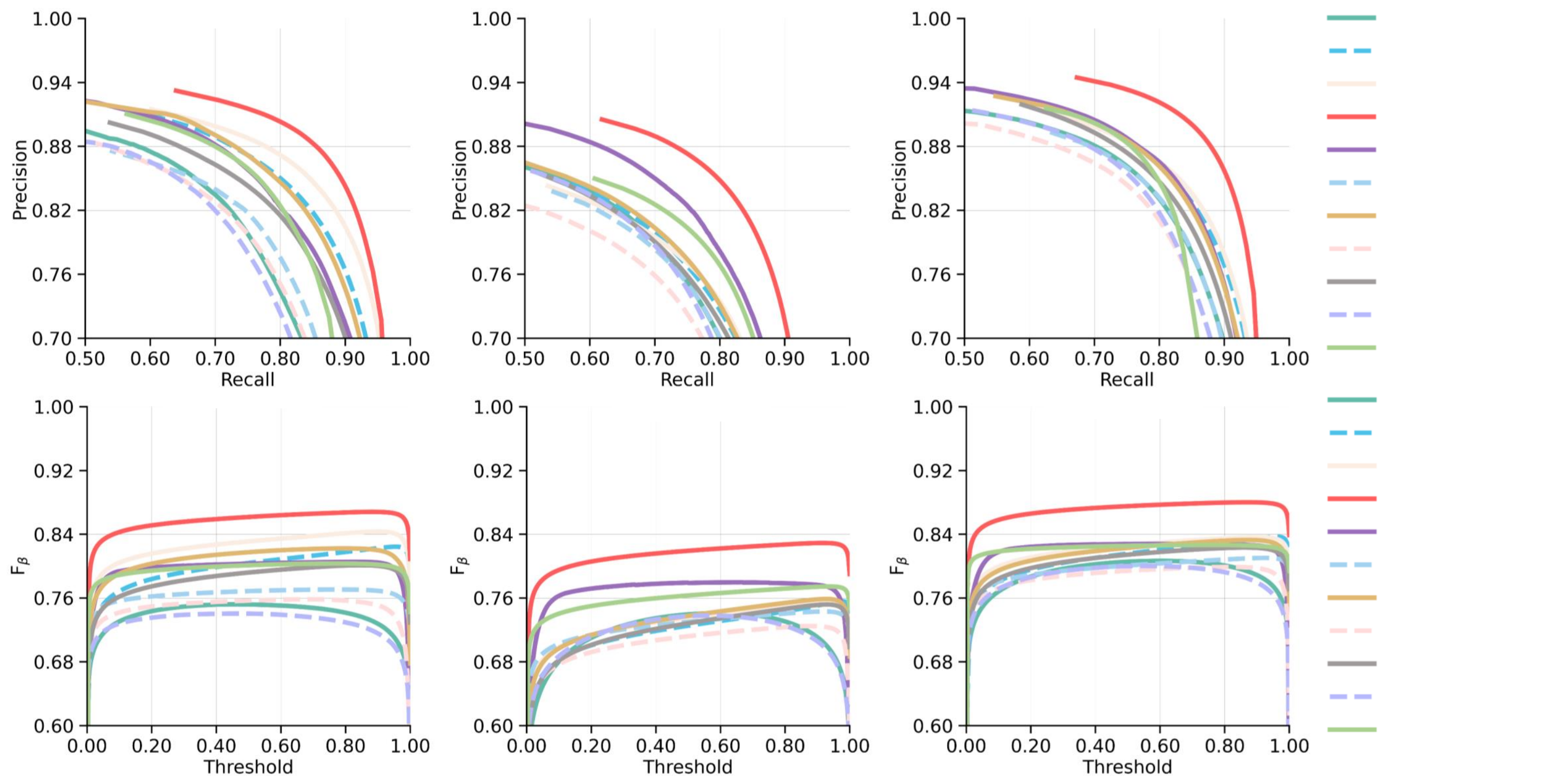}
  \put (-410,227){\footnotesize{CAMO}}
  \put (-410,105){\footnotesize{CAMO}}
  \put (-275,227){\footnotesize{COD10K}}
  \put (-275,105){\footnotesize{COD10K}}
  \put (-135,227){\footnotesize{NC4K}}
  \put (-135,105){\footnotesize{NC4K}}
  \put (-52,229){\footnotesize{UGTR~\cite{yang2021uncertainty}}}
  \put (-52,219){\footnotesize{ICON~\cite{zhuge2022salient}}}
  \put (-52,209){\footnotesize{DTINet~\cite{liu2022boosting}}}
  \put (-52,199){\footnotesize{Ours}}
  \put (-52,189){\footnotesize{ZoomNet~\cite{Pang_2022_CVPR}}}
  \put (-52,179){\footnotesize{C$^2$FNet~\cite{suncontext}}}
  \put (-52,169){\footnotesize{DGNet~\cite{ji2022deep}}}
  \put (-52,159){\footnotesize{PFNet~\cite{mei2021camouflaged}}}
  \put (-52,149){\footnotesize{SINetV2~\cite{fan2022concealed}}}
  \put (-52,139){\footnotesize{MGL-R~\cite{zhai2021mutual}}}
  \put (-52,129){\footnotesize{SegMaR~\cite{Jia_2022_CVPR}}}
  \put (-52,114){\footnotesize{UGTR~\cite{yang2021uncertainty}}}
  \put (-52,104){\footnotesize{ICON~\cite{zhuge2022salient}}}
  \put (-52,94){\footnotesize{DTINet~\cite{liu2022boosting}}}
  \put (-52,84){\footnotesize{Ours}}
  \put (-52,74){\footnotesize{ZoomNet~\cite{Pang_2022_CVPR}}}
  \put (-52,64){\footnotesize{C$^2$FNet~\cite{suncontext}}}
  \put (-52,54){\footnotesize{DGNet~\cite{ji2022deep}}}
  \put (-52,44){\footnotesize{PFNet~\cite{mei2021camouflaged}}}
  \put (-52,34){\footnotesize{SINetV2~\cite{fan2022concealed}}}
  \put (-52,24){\footnotesize{MGL-R~\cite{zhai2021mutual}}}
  \put (-52,14){\footnotesize{SegMaR~\cite{Jia_2022_CVPR}}}
  \vspace{-5pt}
  \caption{PR and $f_{\beta}$ curves of the proposed \nameofmethod{} and the recent SOTA algorithms on three COD datasets.
  }\vspace{-5pt}
  \label{fig:fm}
\end{figure*}

\section{PR \& F$_{\beta}$ curves of COD methods}

We provide the PR and $F_{\beta }$ curves of our CamoFormer and previous methods on NC4K, CAMO, and COD10K datasets, as shown in ~\figref{fig:fm}. Note that the higher the curve is, the better the model performs. It is clear that our CamoFormer (red curve) surpasses all other methods. 

{\small
\bibliographystyle{ieee_fullname}
\bibliography{egbib}
}

\end{document}